\documentclass[runningheads]{llncs}

\usepackage[mobile]{eccv}

\usepackage{eccvabbrv}

\usepackage{graphicx}
\usepackage{booktabs}

\usepackage[accsupp]{axessibility}  % Improves PDF readability for those with disabilities.

\usepackage{multirow}

\usepackage{hyperref}

\usepackage{orcidlink}

\usepackage[dvipsnames]{xcolor}

\DeclareMathOperator*{\argmax}{arg\,max}

\newcommand{\mbf}[1]{\mathbf{#1}}

\newcommand{\getz}[1]{[{#1}]_z}

\newcommand{\Twc}{\mbf{T}_{\text{WC}}}

\newcommand{\Twck}{\mbf{T}_{\text{WC}_r}}
\newcommand{\Rwck}{\mbf{R}_{\text{WC}_r}}
\newcommand{\twck}{\mbf{t}_{\text{WC}_r}}

\newcommand{\Twct}{\mbf{T}_{\text{WC}_t}}
\newcommand{\Rwct}{\mbf{R}_{\text{WC}_t}}
\newcommand{\twct}{\mbf{t}_{\text{WC}_t}}

\newcommand{\Pw}{\mbf{P}_{\text{W}}}
\newcommand{\Pc}{\mbf{P}_{\text{C}}}
\newcommand{\Pck}{\mbf{P}_{\text{C}_r}}
\newcommand{\Pct}{\mbf{P}_{\text{C}_t}}

\newcommand{\pix}{\mbf{p}}

\newcommand{\K}{\mbf{K}}
\newcommand{\dm}{\mbf{d}_{\text{M}}}
\newcommand{\dn}{\mbf{d}_{\text{N}}}
\newcommand{\Kmm}{\mbf{K}_{\text{MM}}}
\newcommand{\Kmminv}{\left( \Kmm \right)^{-1}}
\newcommand{\Knm}{\mbf{K}_{\text{NM}}}
\newcommand{\Kmn}{\mbf{K}_{\text{MN}}}
\newcommand{\Knn}{\mbf{K}_{\text{NN}}}

\newcommand{\Kmmj}{\mbf{K}_{\text{M}_{1:j} \text{M}_{1:j}}}
\newcommand{\Kmminvj}{\left( \Kmmj \right)^{-1}}
\newcommand{\Knmj}{\mbf{K}_{\text{NM}_{1:j}}}
\newcommand{\Kmnj}{\mbf{K}_{\text{M}_{1:j} \text{N}}}

\newcommand{\dmone}{\mbf{d}_{\text{m}_1}}
\newcommand{\dmtwo}{\mbf{d}_{\text{m}_2}}

\newcommand{\jac}[2]{ \frac{\partial{#1}}{\partial{#2}}}

\def\thickhline{\noalign{\hrule height.8pt}}

\begin{document}

% ---------------------------------------------------------------
% TODO REVIEW: Replace with your title
\title{COMO: Compact Mapping and Odometry} 

% TODO REVIEW: If the paper title is too long for the running head, you can set
% an abbreviated paper title here. If not, comment out.
\titlerunning{COMO}

% TODO FINAL: Replace with your author list. 
% Include the authors' OCRID for the camera-ready version, if at all possible.
% \author{Eric Dexheimer\inst{1}\orcidlink{0000-0003-2402-6742} \and
% Andrew J. Davison\inst{1}\orcidlink{0000-0002-3784-099X}}
\author{Eric Dexheimer \and Andrew J. Davison}

% TODO FINAL: Replace with an abbreviated list of authors.
% \authorrunning{F.~Author et al.}
\authorrunning{E.~Dexheimer and A.J.~Davison}
% First names are abbreviated in the running head.
% If there are more than two authors, 'et al.' is used.

% TODO FINAL: Replace with your institution list.
\institute{Dyson Robotics Lab, Imperial College London \\
\email{ \{e.dexheimer21, a.davison\}@imperial.ac.uk}\\
}
% \url{https://www.imperial.ac.uk/}}

\maketitle

\begin{abstract}
  We present COMO, a real-time monocular mapping and odometry system that encodes dense geometry via a compact set of 3D anchor points.  Decoding anchor point projections into dense geometry via per-keyframe depth covariance functions guarantees that depth maps are joined together at visible anchor points.  The representation enables joint optimization of camera poses and dense geometry, intrinsic 3D consistency, and efficient second-order inference.  To maintain a compact yet expressive map, we introduce a frontend that leverages the covariance function for tracking and initializing potentially visually indistinct 3D points across frames.  Altogether, we introduce a real-time system capable of estimating accurate poses and consistent geometry. 
\end{abstract}

%%%%%%%%% BODY
\section{Introduction}
\label{sec:intro}

Achieving accurate and consistent poses and dense geometry from monocular images in real-time is a challenging yet essential undertaking to push forward state-of-the-art in robotics and augmented reality.  Monocular cameras are the key to achieving low-cost, energy-efficient, and compact intelligent platforms. While images contain rich visual information, reconstructing the world is challenging due to the lack of direct geometric observations.

The ideal representation for real-time monocular visual odometry (VO) and simultaneous localization and mapping (SLAM) remains elusive.  While sparse methods jointly optimize camera poses and a set of conditionally independent 3D points given poses, the map lacks consistent dense geometry for downstream tasks and pose estimation cannot benefit from all visual information.  Dense SLAM uses and reconstruct all pixels, but the sheer number of variables relative to measurements renders joint optimization infeasible and inference ill-posed.  Recently, learned priors over compact depth map representations, such as codes and sparse 2D depths, have enabled joint optimization.  However, representations with 2D depth priors have thus far lagged in accuracy, as consistency between depth maps in 3D is not guaranteed.  Volumetric representations, such as voxel grids and neural fields, directly enforce dense 3D consistency by construction, but cannot perform real-time joint optimization due to expensive rendering.  Furthermore, compact, efficient, and expressive 3D priors for general scenes are not yet suitable.  Achieving accurate and consistent 3D geometry in real-time would represent a significant advance for monocular SLAM.

\begin{figure*}[t]
  \begin{subfigure}{0.48\textwidth}
    % \captionsetup{justification=centering}
    \includegraphics[width=\linewidth]{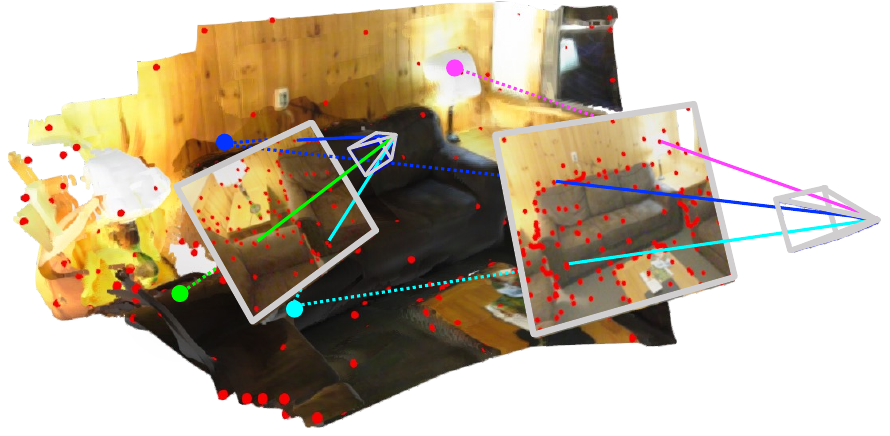}
    \caption{Point cloud reconstruction from the sliding-window monocular odometry defined by 240 points and 8 keyframes.} 
    \label{fig:title_fig_a}
  \end{subfigure}%
  \hfill
  % \hspace*{\fill}   % maximize separation between the subfigures
  \begin{subfigure}{0.48\textwidth}
    % \captionsetup{justification=centering}
    \includegraphics[width=\linewidth]{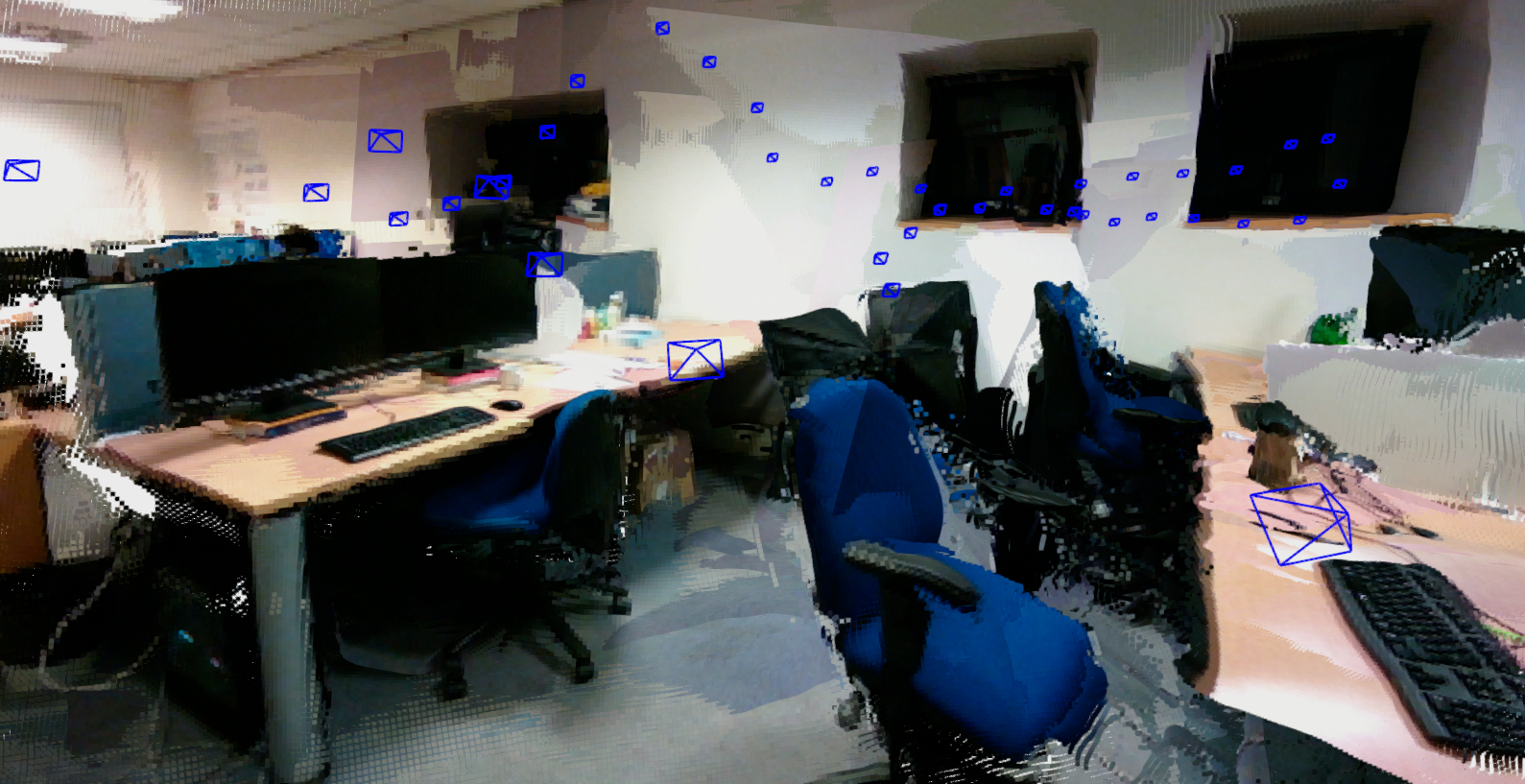}
    \caption{Dense point cloud from 40 keyframes across a large office during live operation of a monocular camera.}     
    \label{fig:title_fig_b}
  \end{subfigure}%

  \caption{COMO encodes scene geometry via a compact set of 3D anchor points and decodes dense geometry via per-keyframe depth covariance functions.  The 3D points visualized in red anchor depth maps together from multiple views while the covariance function generates dense geometry by conditioning on sparse point projections.} 
  \label{fig:title_fig}
\end{figure*}

In this work, we propose a representation that achieves the three desired properties: joint optimization of poses and dense geometry, intrinsic 3D consistency, and low-latency, real-time operation.  We encode the scene as a compact set of 3D anchor points, which are shared across frames.  To decode into dense geometry, 3D points are first projected into keyframes. We then use image-conditioned depth covariance functions \cite{dexheimer_learning_2023} per keyframe to generate dense depth maps that intersect each visible 3D point.  Depth maps are backprojected into 3D for calculating any-pixel photometric error.  All steps are efficient and have analytic Jacobians for second-order optimization.  Updates to dense geometry are propagated into our encoded scene representation and ensure suitable regularization for the ill-posed problem of dense reconstruction.

Maintaining a compact map in an incremental VO setting requires a compatible frontend.  Keeping a small number of points benefits efficiency, while the distribution of points affects expressiveness.  Feature-based systems focus on visually distinct features, but this often leads to many points on edges which are not suitable for modeling dense geometry.  Therefore, we leverage depth covariance for determining anchor point visibility in new keyframes, actively initializing new 3D points, and encoding the current dense geometry into new 3D points.

Our system demonstrates improved robustness over methods lacking joint pose and dense geometry optimization, as well as more accurate pose and geometry estimation compared to compact 2D representations.  Altogether, we propose a real-time VO system that achieves robust and accurate odometry along with consistent dense geometry as shown in \cref{fig:title_fig}.

In summary, the contributions of our work are:
\begin{itemize}
	\item An efficient representation of dense geometry encoded by a compact set of 3D anchor points and decoded by depth covariance functions.
	\item A frontend for the compact map that leverages depth covariance for visibility, active initialization, and encoding dense geometry.
	\item A real-time monocular visual odometry and mapping system that produces accurate and consistent poses and dense geometry.
\end{itemize}

\section{Related Work}

\noindent \textbf{Sparse vs. Dense Visual SLAM} 
Sparse SLAM systems optimize 3D points that are conditionally independent given poses. Exploiting the sparsity inherent in the information form of structure-from-motion is the key to real-time algorithms with a significant number of points \cite{klein_parallel_2007}.  Modern sparse VO methods \cite{mur-artal_orb-slam2_2017, forster_svo_2017, engel_dso_2018} and large-scale bundle adjustment \cite{schonberger_structure-from-motion_2016} rely on the Schur complement which greatly reduces optimization complexity.  Sparse systems lack dense reconstruction, which is useful for many downstream tasks.  Thus, it is common to first perform pose estimation followed by dense mapping given poses and potentially sparse landmark estimates \cite{newcombe_live_2010, pizzoli_icra_2014, mur-artal_probabilistic_2015, matsuki_codemapping_2021}.  MVS approaches estimate depth maps via a reference cost volume and multiple supporting frames, and have shown progress when integrated into a learning pipeline \cite{koestler_tandem_2022}.  However, the sparse followed by dense paradigm has two major issues.  Dense information is ignored for pose estimation, which can result in failure in many real-world scenarios, while the mapping step is heavily reliant on accurate pose estimation.  Joint optimization of poses and dense geometry avoids the brittle two-step process and has the potential to achieve more accurate odometry and mapping \cite{platinsky_monocular_2017}.

\begin{figure*}[t]
  \begin{subfigure}{0.32\textwidth}
    \captionsetup{justification=centering}
    \frame{\includegraphics[width=\linewidth]{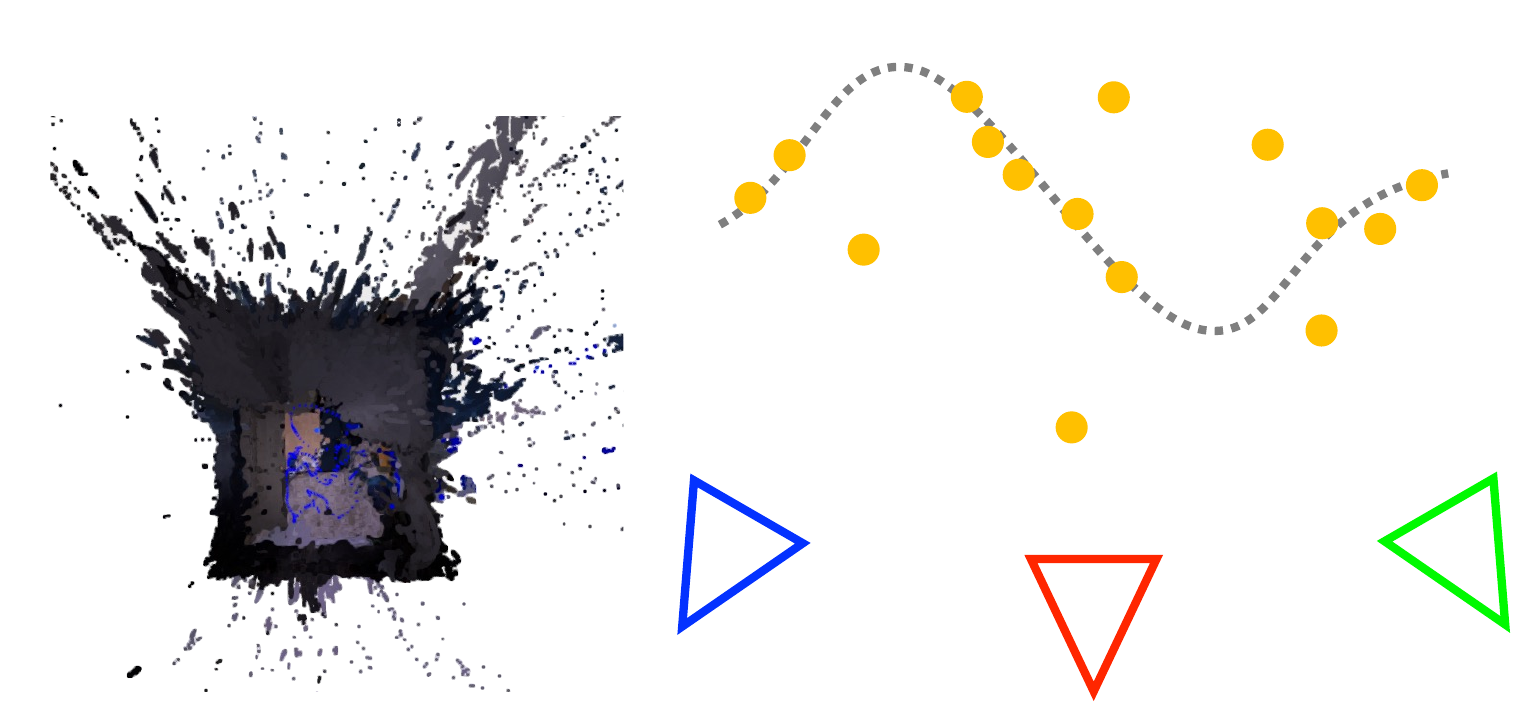}}
    \caption{Dense bundle adjustment with 3D landmarks \cite{teed_droid_2021}} \label{fig:droid_map}
  \end{subfigure}%
  \hfill
  % \hspace*{\fill}   % maximize separation between the subfigures
  \begin{subfigure}{0.32\textwidth}
    \captionsetup{justification=centering}
    \frame{\includegraphics[width=\linewidth]{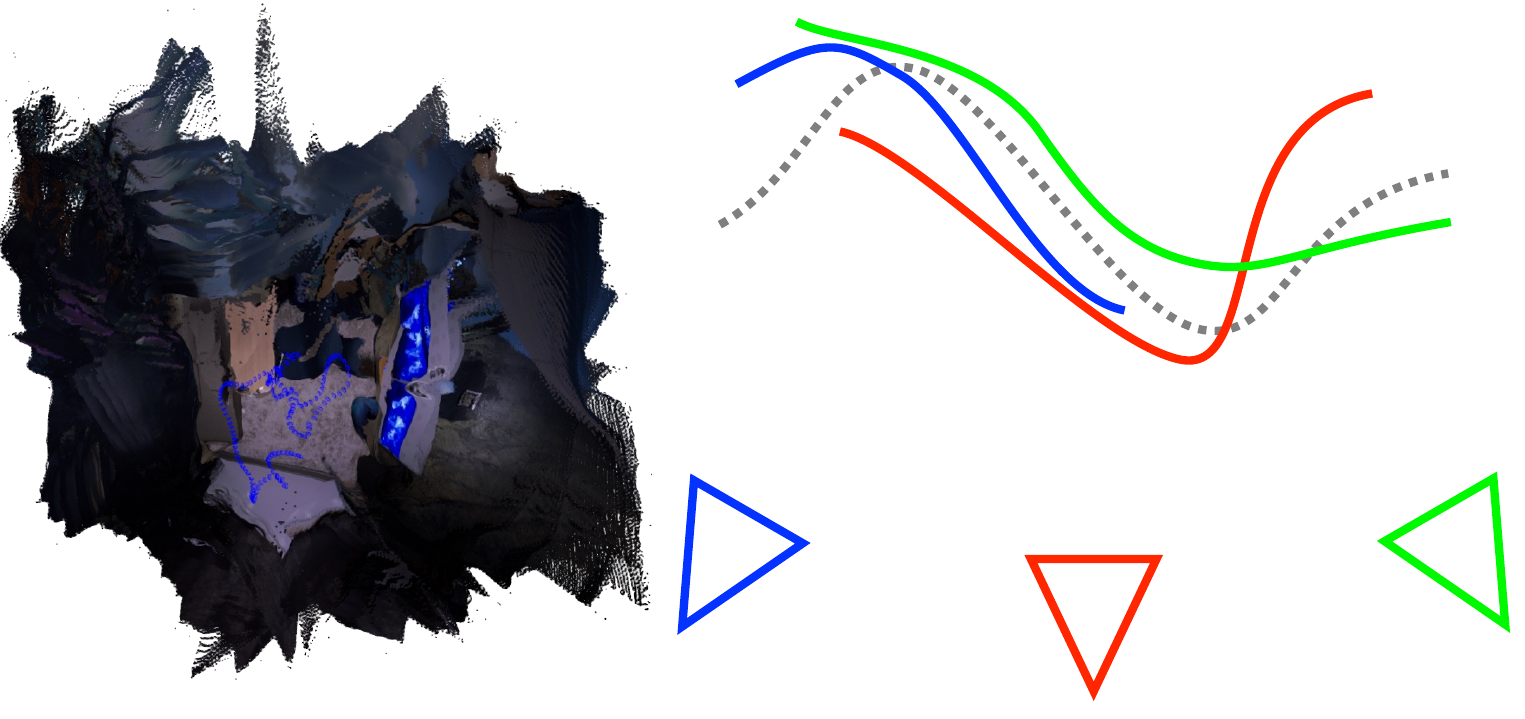}}
    \caption{Per-frame dense depth priors via compact codes \cite{czarnowski_deepfactors_2020}} \label{fig:code_map}
  \end{subfigure}%
  \hfill
  % \hspace*{\fill}   % maximizeseparation between the subfigures
  \begin{subfigure}{0.32\textwidth}
    \captionsetup{justification=centering}
    \frame{\includegraphics[width=\linewidth]{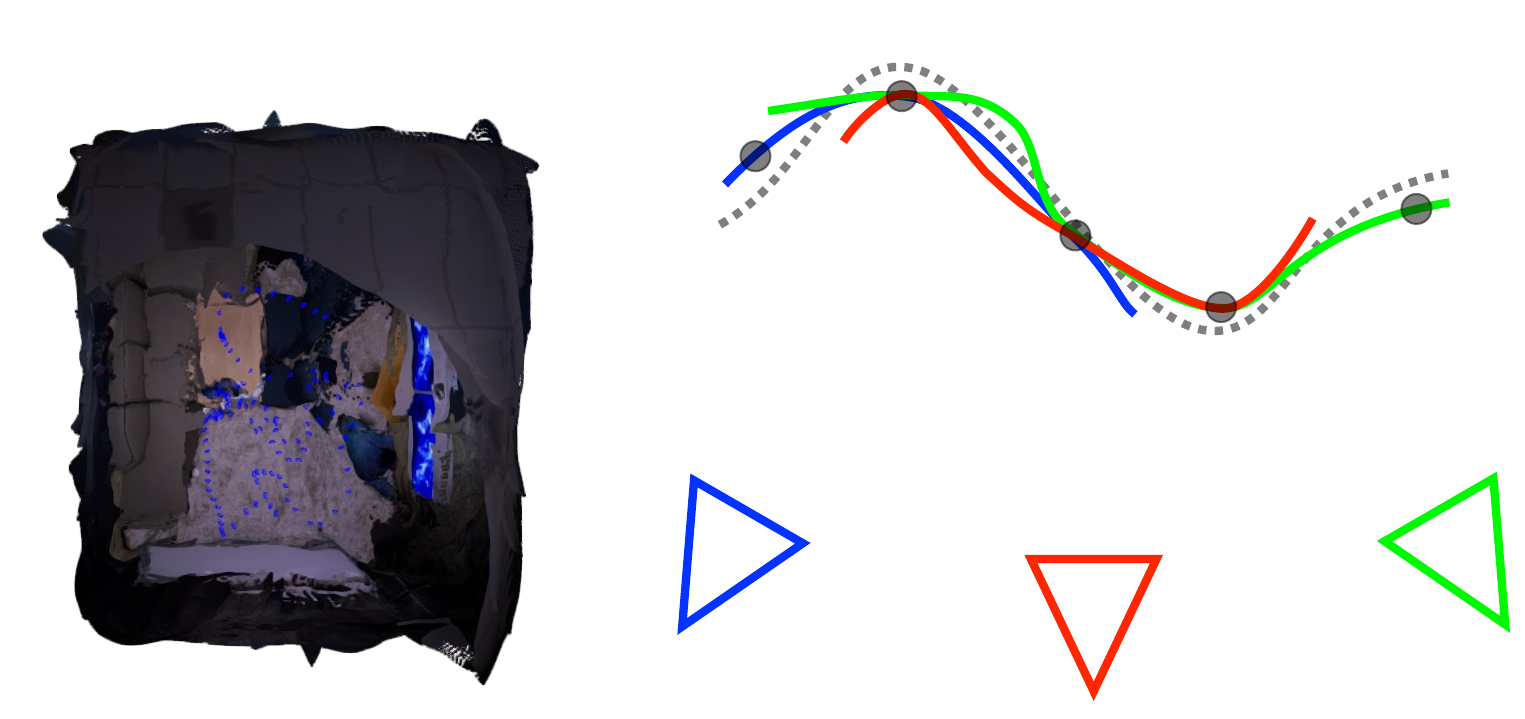}}
    \caption{Ours: 3D points and depth covariance functions} \label{fig:como_map}
  \end{subfigure}
  
    \caption{Reconstructions on Replica and geometry properties of different dense map representations. The dotted line represents the true surface.  (a) Densely reconstructing a large number of conditionally independent 3D points given poses enables accurate pose estimation and many accurate points, but there is no guarantee of coherent geometry.  (b) Depth priors produce smooth depth maps, but even with inter-frame consistency losses, can produce inconsistent geometry and bias preventing global consistency.  (c) A compact set of 3D points and depth covariance functions anchor depth maps together, leading to consistent pose estimation and dense geometry.} 
    \label{fig:maps_3d}
\end{figure*}

Dense methods \cite{newcombe_dtam_2011, engel_lsd-slam_2014} brought the promise of estimating poses and complete geometry simultaneously.  However, joint optimization proved difficult due to the ill-posed nature of reconstructing all points. Hand-crafted geometry priors regularize the problem, but disrupt the sparsity and thus the tractability of joint optimization as discussed in \cite{engel_dso_2018}.  In practice, these methods resort to alternating pose and geometry estimation or move away from second-order optimization.  More recently, DROID-SLAM \cite{teed_droid_2021} reconstructs depths for all pixels, but similar to sparse systems, does so with no geometric correlation.  This results in highly parallelized bundle adjustment, but results in many noisy points and no guarantee of consistent depths within a frame and across frames, as shown in \cref{fig:droid_map}.  All real-time state-of-the-art SLAM systems \cite{campos_orbslam3_2021, engel_dso_2018, teed_droid_2021, teed_dpvo_2023} still rely on the Schur complement, and dense methods with geometric correlation have traditionally not demonstrated comparable pose accuracy, which also limits geometry estimation.

\noindent \textbf{Learned Depth Priors for Compact Optimization}
To achieve consistent poses and geometry, monocular depth priors in SLAM focus on test-time optimization to avoid irreparable errors from single-view depth prediction.  Therefore, depth priors in monocular SLAM often predict a subspace of potential depths given an image via a compact latent code \cite{bloesch_codeslam_2018, czarnowski_deepfactors_2020, matsuki_codemapping_2021}, or a depth map basis \cite{tang_ba-net_2019, graham_ridgesfm_2020}.  Even with this flexibility, depth maps are not guaranteed to be consistent across multiple views since latent variables live only in frames as shown in \cref{fig:code_map}.  While per-pixel depth and keypoint reprojection losses may encourage 3D consistency \cite{czarnowski_deepfactors_2020}, this may introduce bias and does not scale computationally since it requires per-pixel pairwise frame comparisons.  Most importantly, it lacks intrinsic 3D consistency, and depth maps often produce a layering effect in 3D as in \cref{fig:code_map}. Predicting a view-based mesh topology permits optimizing depth vertices \cite{bloesch_learning_2019}, but it is unclear how to move to a 3D topology.

In this work, we leverage the recently proposed learned depth covariance function \cite{dexheimer_learning_2023}. While \cite{dexheimer_learning_2023} introduces an odometry formulation, it only optimizes per-frame 2D inducing depths, which has the same limitations as per-frame codes, and lacks intrinsic 3D consistency.  Our key insight is that depth covariance permits sharing latent 3D points across frames, so that our representation lives in 3D rather than separate 2D image planes.  While codes and bases are a fixed subspace that cannot guarantee a depth map passes through any set of 3D points, our covariance formulation guarantees that depth maps go through any desired 3D points as shown in \cref{fig:como_map}.  Therefore, we can anchor depth maps together by construction for dense 3D consistency.  We also exploit the covariance function for our frontend in tracking and initializing potentially non-visually distinct points.

\noindent \textbf{Keyframe-based vs. Volumetric Maps}  
Keyframe-based maps host dense depths in camera frames, which maintain flexible level-of-detail rooted in the camera resolution and focus only on surfaces.  However, representing space using per-pixel depths cannot guarantee that inter-frame correspondences refer to the same 3D point.  While pairwise depth constraints can mitigate inconsistency \cite{czarnowski_deepfactors_2020}, these exhaustive per-pixel errors are expensive to compute, lack intrinsic consistency, and may introduce bias when balancing with other losses.  Methods that project hosted depths into neighboring frames and predict flow updates \cite{teed_droid_2021} are asymmetric with no guarantee of cycle consistency.

Volumetric maps live in 3D space and unknown quantities are shared across frames.  3D consistency is achieved by construction but it is challenging to balance fidelity and efficiency.  Voxel grids using backward sensor models \cite{newcombe_kinectfusion_2011, hornung_octomap_2013} assume known poses and independence between cells, and thus are unsuitable for efficient and consistent optimization.  While forward models in voxel grids \cite{thrun_learning_2001}, neural fields \cite{mildenhall_nerf_2020, sucar_imap_2021}, and 3D Gaussian splatting \cite{kerbl_3d_2023, matsuki_gaussian_2024} permit consistent geometry and pose optimization, these representations are limited in low-latency monocular VO and SLAM due to expensive rendering, a large number of correlated variables per ray, and alternating and first-order optimization.  Volumetric methods are also subject to a resolution trade-off.  Higher resolution permits greater detail at the expense of memory, runtime complexity, and higher levels of noise.

Our 3D representation achieves the benefits of both frame-based and volumetric maps for odometry and mapping: the per-pixel resolution and efficiency of depth maps with the intrinsic 3D consistency of volumetric representations.  The efficient compact-to-dense model permits joint optimization of poses and geometry via any-pixel constraints in real-time.

\section{Compact Mapping Backend}
\label{sec:backend}

The mapping backend maintains a compact set of 3D anchor points along with keyframes hosting per-pixel covariance function parameters.  A dense world-centric point cloud is decoded from anchor points and poses, which is refined by second-order minimization of photometric error.  VO operates in a sliding-window fashion with a fixed number of keyframes. \cref{fig:mapping} provides an overview.

Quantities in the world frame and camera frame, such as points, are denoted by subscripts $\Pw$ and $\Pc$, respectively.  Anchor points are indexed by superscripts \textit{m}, while dense points are indexed \textit{n}, such as pixels $\pix^m$ and $\pix^n$.  Stacked vector forms of sparse and dense variables use capital M and N, as in log-depth vectors $\dm$ and $\dn$.  The notation $||\mbf{r}||^2_\mathrm{\Sigma}$ denotes Mahalanobis distance of residual vector $\mbf{r}$ with measurement covariance matrix $\mathrm{\Sigma}$.

\subsection{Preliminaries: Depth Covariance Function}
\label{subsec:depth_cov}

First, we introduce the depth covariance function from \cite{dexheimer_learning_2023}.  This covariance function models the distribution over all possible log-depth functions for any finite set of pixels via a Gaussian process (GP) \cite{rasmussen_gaussian_2005}.  First, a CNN takes in an RGB image and outputs a per-pixel feature map $\phi$.  Compared to \cite{dexheimer_learning_2023}, we use a zero-mean GP prior for simplicity, such that the distribution is defined only by the covariance function $k$ that takes in two pixels and respective CNN features:
\begin{align}
    d(\mbf{p}) \sim \mathcal{GP}\left(0,\, k \left( [\pix; \phi(\pix)], [\pix'; \phi(\pix')] \right) \right) .
\end{align}
We found that the mean log-depth variable proposed in \cite{dexheimer_learning_2023} is unnecessary, as the largest eigenvector of the covariance function corresponds to scale of the entire depth map. More information on the exact covariance function $k$ is detailed in \cite{dexheimer_learning_2023}, but the key idea is that larger covariance is an indicator of similarity in log-depth.  Since the covariance function takes in both pixel locations and CNN features, it is nonstationary which permits varying levels of smoothness and discontinuities conditioned on the image content.  The zero-mean prior defines a planar depth prior in the absence of observations.  To define a covariance matrix $\K$ for a set of pixels, the $(i^\text{th}, j^\text{th})$ entry of $\K$ is filled in by $k \left( [\pix^i; \phi(\pix^i)], [\pix^j; \phi(\pix^j)] \right)$.

\subsection{Compact-to-Dense Geometry}
\label{subsec:s2d}

\begin{figure*}[t]
  \begin{subfigure}{0.23\textwidth}
    \captionsetup{justification=centering}
    \includegraphics[width=\linewidth]{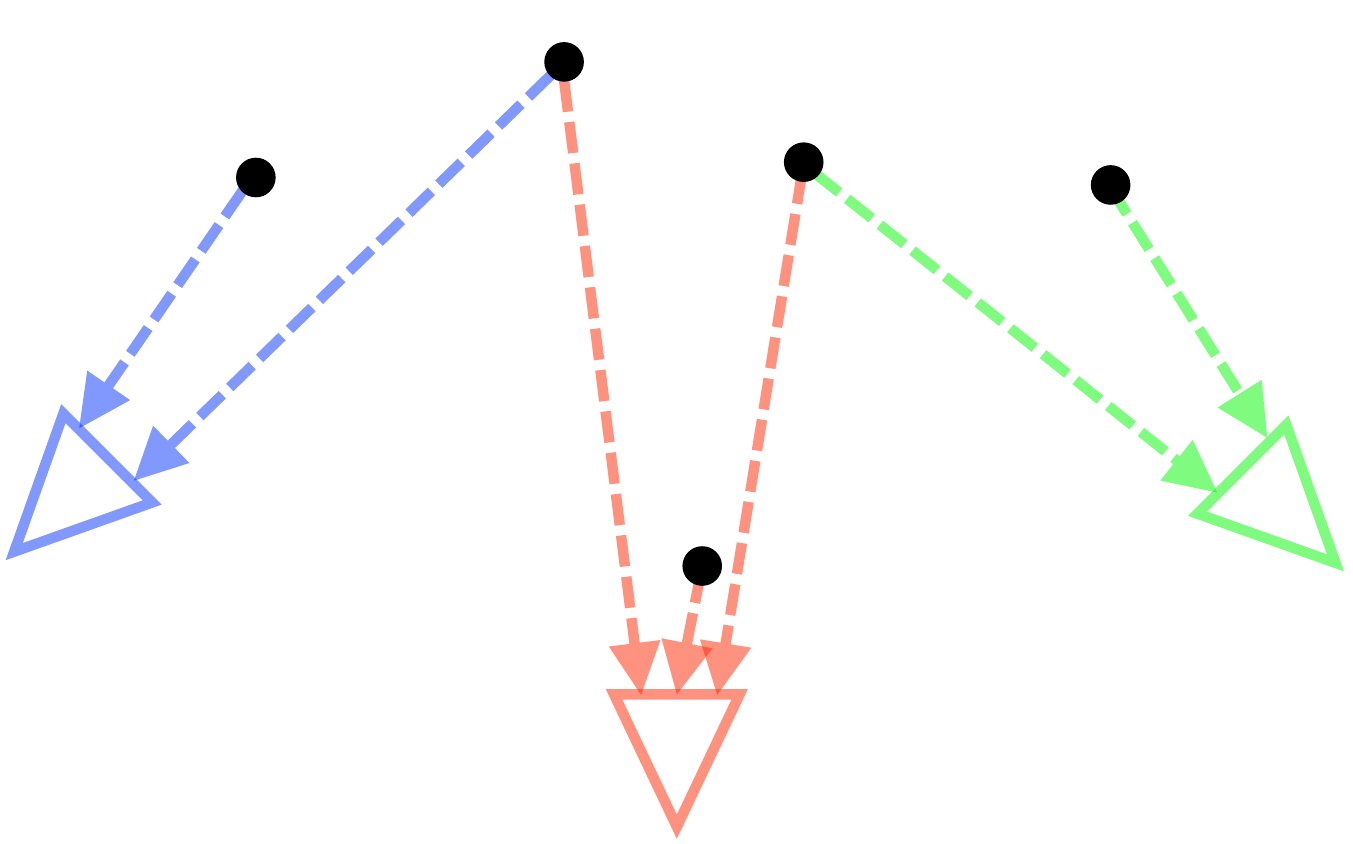}
    \caption{Anchor point projection} \label{fig:mapping_a}
  \end{subfigure}%
  \hfill
  % \hspace*{\fill}   % maximize separation between the subfigures
  \begin{subfigure}{0.23\textwidth}
    \captionsetup{justification=centering}
    \includegraphics[width=\linewidth]{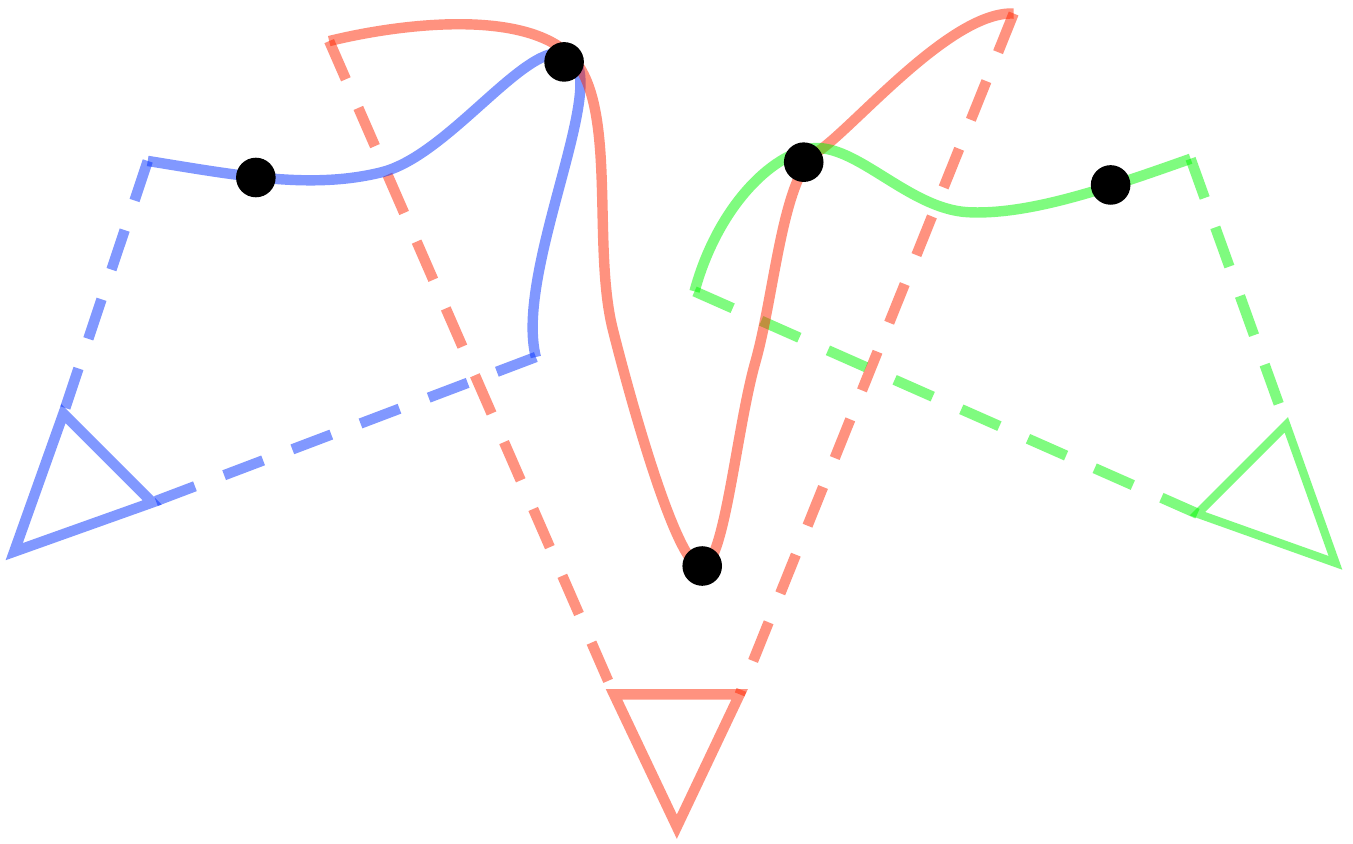}
    \caption{Dense geometry construction} \label{fig:mapping_b}
  \end{subfigure}%
  \hfill
  % \hspace*{\fill}   % maximizeseparation between the subfigures
  \begin{subfigure}{0.23\textwidth}
    \captionsetup{justification=centering}
    \includegraphics[width=\linewidth]{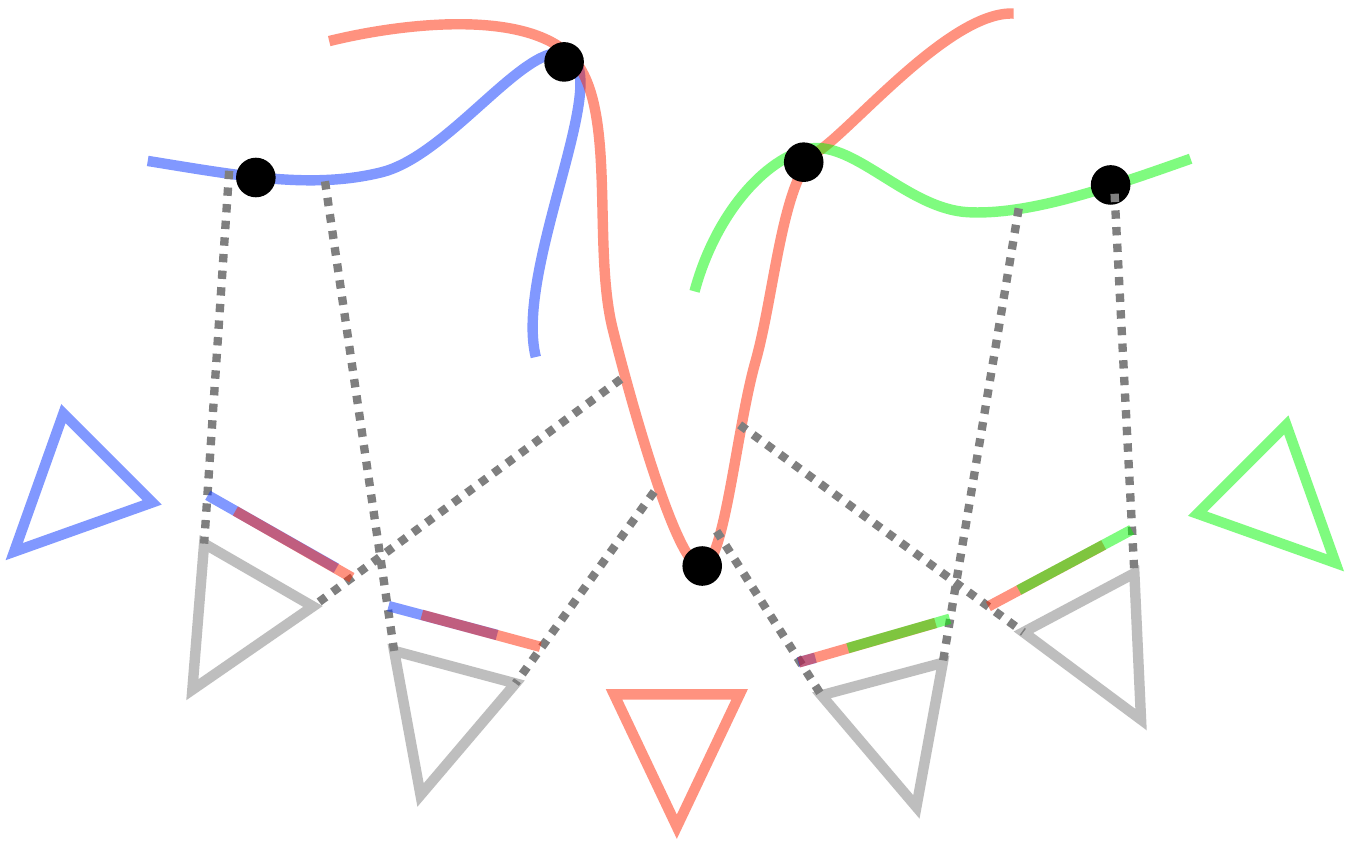}
    \caption{Photometric error calculation} \label{fig:mapping_c}
  \end{subfigure}
  \hfill
  % \hspace*{\fill}   % maximizeseparation between the subfigures
  \begin{subfigure}{0.23\textwidth}
    \captionsetup{justification=centering}
    \includegraphics[width=\linewidth]{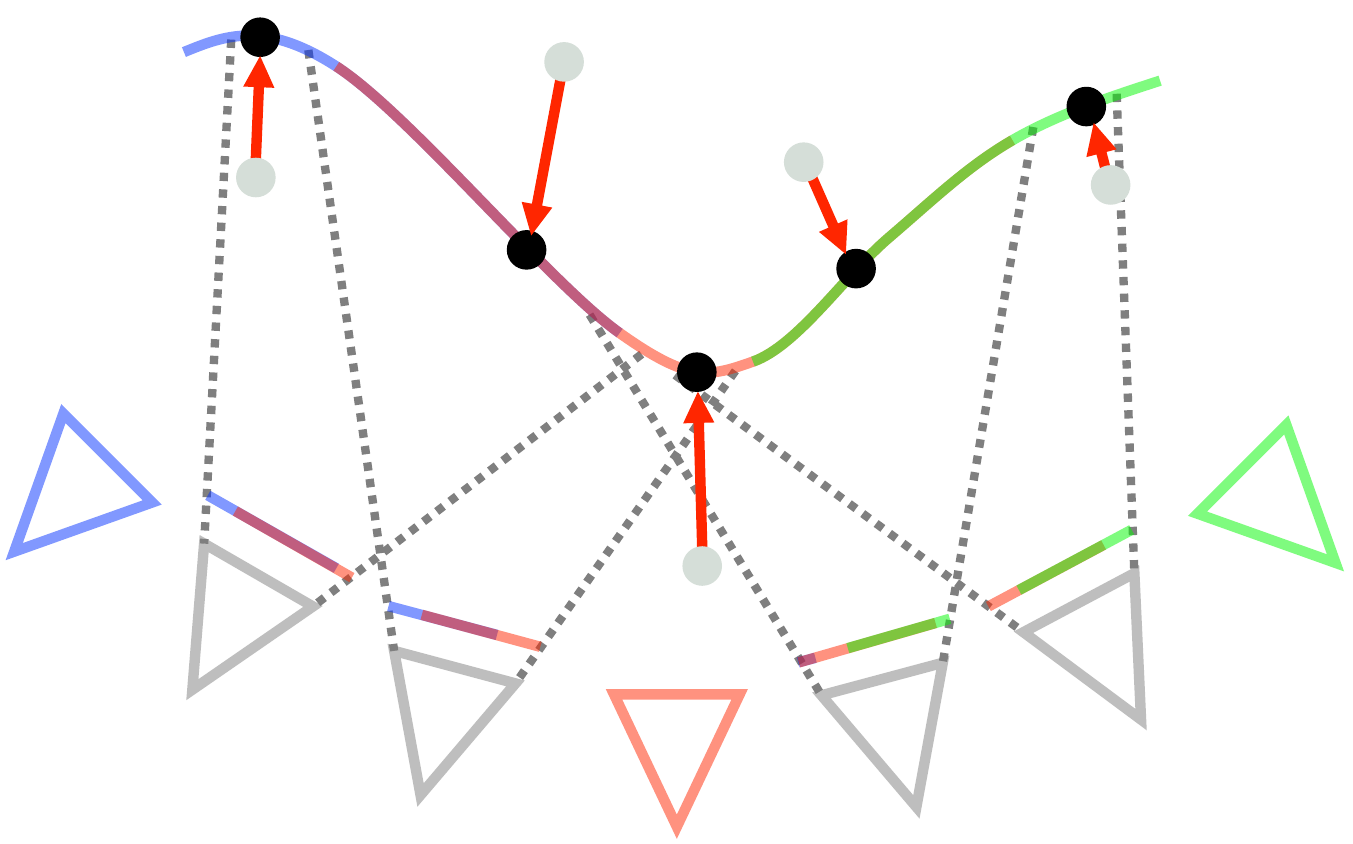}
    \caption{Pose and geometry update} \label{fig:mapping_d}
  \end{subfigure}

  \caption{Overview of compact mapping framework. (a) Anchor points are projected to visible keyframes.  (b) Keyframes decode into dense depth and backproject geometry to 3D.  (c) Target frames enforce dense photo-consistency.  (d) Optimizing dense alignment error leads to updated poses and geometry with greater 3D consistency.} 
  \label{fig:mapping}
\end{figure*}

Given anchor point to keyframe visibility, we generate a dense depth map given anchor points $\Pw^m$, the pose of keyframe $r$ with rotation $\Rwck$ and translation $\twck$, and the keyframe image.  Anchor points are transformed and projected to pixel coordinates $\pix$ via camera projection $\pi$:
\begin{align}
    \Pck^m = \Rwck^T (\Pw^m - \twck), \qquad \pix^m = \pi(\Pck^m),
\end{align}
which is also illustrated in \cref{fig:mapping_a}.  At this point, we create a dense log-depth map from log-depth observations  $d^m \triangleq \log{ \left(\getz{\Pck^m} \right) }$ and projected pixel coordinates $\pix^m$.  Given an image and the covariance function $k$ for any two pixels from \cref{subsec:depth_cov}, we can construct the covariance matrix between all visible anchor point pixels in $\Kmm$, and the covariance between all image pixels and anchor point pixels in $\Knm$. We also stack all training $d^m$ and test $d^n$ log-depths into $\dm$ and $\dn$, respectively.  The GP formulation allows efficient dense log-depth prediction via the linear Gaussian conditioning equation:
\begin{align}
    \label{eq:gp_cond_mean}
    \dn &= \Knm \Kmminv \dm.
\end{align}
The GP guarantees that each dense log-depth map goes through the anchor points, creating a consistent surface across multiple views.  Next, we backproject points for each pixel and transform to world coordinates:
\begin{align}
    \Pw^n = \Rwck \pi^{-1}(\pix^n, e^{d^n}) + \twck,
\end{align}
where $n$ indexes a query pixel.  Note that each keyframe generates a dense depth map via its observed anchor points and covariance parameter feature maps as in \cref{fig:mapping_b}.  This compact-to-dense formulation only requires transformations, camera projection, and linear GP conditioning.  Therefore, we can efficiently calculate analytical Jacobians for dense per-pixel constraints with respect to poses and anchor points.  We show the chain rule for these steps in Section A.1.

In practice, we ignore the Jacobians with respect to pixel coordinates when constructing $\Kmm$ and $\Knm$ and calculate the covariance matrices once at initialization.  In addition to being more efficient, this ensures that depth maps do not undergo unstable changes if a point moves across a depth discontinuity.  If a visible anchor point moves behind a camera during optimization, we reinitialize it to the median depth of the keyframe.

\subsection{Photometric Residuals}

While only keyframes are used to create dense geometry, additional small baseline support frames are included to improve convergence.  We define a set of edges $\mathcal{E}$ for photometric error that consists of temporally adjacent keyframe-keyframe pairs, and support frames to the two nearest keyframes in time.  Given a target image $I_t$, which may be either a keyframe or support frame, with pose $\Twct$, we first transform world points from keyframe view $r$ and project:
\begin{align}
    \pix^n_t &= \pi(\Rwct^T (\Pw^n - \twct)).
\end{align}
Given these projective correspondences and accounting for exposure and global illumination changes via affine brightness parameters, $a$ and $b$, similar to \cite{engel_dso_2018}, we calculate the photometric error across pixels and frame edges:
\begin{align}
    E = \sum_{r,t \in \mathcal{E}} \sum_n || \mathbf{r}^n_{r,t} || ^2_{\sigma_\mbf{r}^2 I}, \qquad \qquad
    \mathbf{r}^n_{r,t} = I_t(\mbf{p}^n_t) + b_t - \left( \frac{e^{-a_r}}{e^{-a_t}} I_r(\mbf{p}^n_r) + b_r \right),
\end{align}
which is illustrated in \cref{fig:mapping_c}. These constraints depend on each frame's pose and affine parameters, and the anchor points viewed by the reference keyframe.  For robustness, we perform iteratively reweighted least squares (IRLS)  with a Huber cost function and set the photometric standard deviation $\sigma_\mbf{r}$ to be proportional to the median absolute residual \cite{alismail_direct_2016} to handle diverse scenes with different error scales.  Compared to RGB residuals, we found that grayscale images were more efficient and similarly robust.  The Jacobian chain rule for photometric residuals is in Section A.2.

\subsection{Additional Constraints}

While photometric error is the only data term, we introduce additional priors to remove ambiguities.  Priors on the oldest keyframe's pose and affine brightness parameters remove gauge freedom in the sliding-window.  After removing a keyframe from the window, we include priors on its still-observed landmarks to ensure global consistency as a simpler alternative to first-estimate Jacobians \cite{engel_dso_2018}.

Since keyframes may have regions unobserved by other frames, which causes anchor points to lack photometric constraints, we include two geometry priors.  The first is a weak prior on anchor points to match the log median depth $s$ of the keyframe it was first observed in.  The second is a GP prior $||\dm - s||_{\Kmm}^2$ on every keyframe to regularize anchor points that have large correlation, which gives improved estimates for underconstrained anchor points lying on surfaces with well-constrained ones, such as on textureless walls.

During optimization, there is ambiguity in whether an anchor point changes depth or moves laterally along a surface, such as when viewing a plane at an angle. Thus, we include a pixel prior encouraging the projection of the point in its initializing keyframe not to change from its first observation, which defines a ray-surface intersection.  As mentioned previously, this is more efficient than calculating Jacobians through \cref{subsec:s2d} and more stable since GP inducing point optimization has little benefit when points start in suitable locations \cite{burt_convergence_2020} as ensured by our frontend in \cref{sec:frontend}.

\subsection{Second-Order Optimization}

We calculate analytical Jacobians for all constraints in the optimization.  The compactness of our linear system permits real-time joint optimization of poses and correlated geometry that is infeasible for traditional dense VO.  Stacking all Jacobians into $\mbf{J}$, weights from IRLS and noise models into the diagonal matrix $\mbf{W}$, and residuals into $\mbf{r}$, we solve for parameter updates $\Delta \mbf{x}$ shown in \cref{fig:mapping_d} via dense Cholesky factorization of the Gauss-Newton normal equations:
\begin{align}
    (\mbf{J}^T \mbf{W} \mbf{J}) \Delta \mbf{x} = -\mbf{J}^T \mbf{W} \mbf{r} .
\end{align}
Pose updates are minimally parameterized as Lie-algebra elements $\mathfrak{se}(3)$ as is standard in SLAM \cite{sola_micro_2018}.  We found that the most time consuming step of backend optimization is accumulating geometry blocks into the Hessian $(\mbf{J}^T \mbf{W} \mbf{J})$.  This can be sped up by factoring out constant components of the Jacobians and reducing the dimensionality of the accumulation as detailed in Section A.4.

\section{Visual Frontend}
\label{sec:frontend}

\begin{figure*}[t]
  \begin{subfigure}{0.23\textwidth}
    \captionsetup{justification=centering}
    \includegraphics[width=\linewidth]{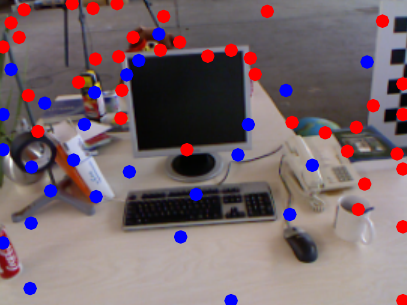}
    \caption{Anchor points visible in KF \#1} \label{fig:corr_a}
  \end{subfigure}%
  \hfill
  % \hspace*{\fill}   % maximize separation between the subfigures
  \begin{subfigure}{0.23\textwidth}
    \captionsetup{justification=centering}
    \frame{\includegraphics[width=\linewidth]{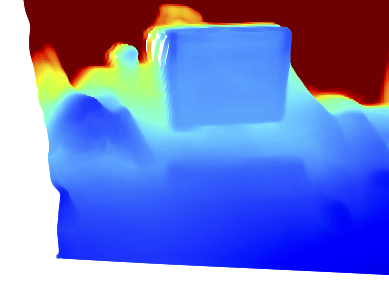}}
    \caption{KF \#1 depths projected to KF \#2} \label{fig:corr_b}
  \end{subfigure}%
  \hfill
  % \hspace*{\fill}   % maximizeseparation between the subfigures
  \begin{subfigure}{0.23\textwidth}
    \captionsetup{justification=centering}
    \includegraphics[width=\linewidth]{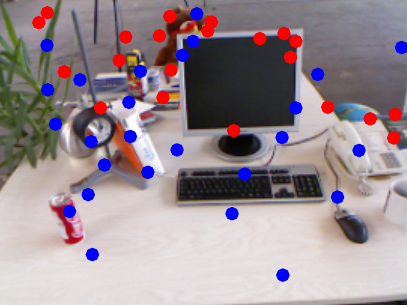}
    \caption{Anchor point matches in KF \#2} \label{fig:corr_c}
  \end{subfigure}
  \hfill
  % \hspace*{\fill}   % maximizeseparation between the subfigures
  \begin{subfigure}{0.23\textwidth}
    \captionsetup{justification=centering}
    \includegraphics[width=\linewidth]{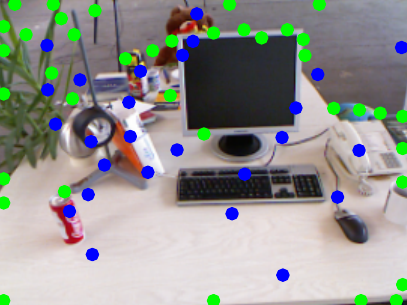}
    \caption{Newly initialized points in KF \#2} \label{fig:corr_d}
  \end{subfigure}

  \caption{Overview of visibility checks between two keyframes (KF).  Matches are shown in blue, rejected matches in red, and newly initialized points in green.  Note that occluded edges are rejected, while non-visually distinct points are often matched.  New points are allocated to geometrically complex regions while the table is already well-represented.} \label{fig:corr}
\end{figure*}

The visual frontend tracks the current frame with respect to the map, selects keyframes and support frames for the backend, determines 3D anchor point visibility in new keyframes, and initializes new 3D anchor points.  Compared to feature-based systems, the ideal points for an expressive yet compact map may be visually indistinct, so we utilize depth covariance for robust visibility checks. 

\subsection{Dense Photometric Tracking}

Given the current estimate of the newest keyframe's depth, we optimize the pose and affine brightness parameters using grayscale image alignment in the IRLS framework \cite{baker_lucas_2004}.  For efficiency, we perform inverse compositional tracking and use a multi-scale image pyramid.  After tracking, a new keyframe is added to the backend if there is either significant translation relative to the median scene depth or if the number of unique projected pixels in the tracked frame falls below a threshold to handle rotation and occlusion.  Support frames are added via modified thresholds according to the desired frequency between keyframes.

\subsection{Anchor Point Visibility}

When a new keyframe is added, we perform explicit visibility checks to 3D anchor points viewed in the last keyframe to achieve 3D consistency.  Note that we only check for visibility as anchor points have no explicit constraint on its location in the image.  Compared to feature-based methods \cite{mur-artal_orb-slam2_2017}, anchor points often lack visual distinctiveness, so matching via image information is problematic. Instead, we leverage the interpretability of the depth covariance function to determine visibility.  In general, an anchor point should be compatible with all depth covariance functions that use it to generate dense geometry. 

First, the 3D anchor points and dense depth map from the last keyframe, shown in \cref{fig:corr_a}, are projected into the current frame, shown in \cref{fig:corr_b}.  Given the projected sparse and dense pixel coordinates, we may query covariance parameters $\phi(\mbf{p})$ and construct $\Kmm$ and $\Kmn$ for the new keyframe.  We leverage the linearity of the GP to compress dense log-depth observations into the sparse projected anchor point coordinates.  Assuming conditional independence of log-depth observations $\mbf{d}_n$ given the sparse log-depths $\mbf{d}_m$, we can solve an efficient least-squares problem: 
\begin{align}
    \label{eq:compress_depth}
    \min_{\dm}  || \Knm \Kmminv \dm  - \dn ||_{\sigma_d^2} ^2 + ||\dm ||_{\Kmm}^2,
\end{align}
where the first conditional term ensures that $\dm$ fits the projected dense observations with standard deviation $\sigma_d$, while the second prior term regularizes sparse log-depths according to the GP training covariance $\Kmm$.  We perform a consistency check on the log-depth difference between the current 3D anchor points and those solved in \cref{eq:compress_depth}, and prune points at depth discontinuities.  This yields potential matches the are consistent between the two keyframes.  

In practice, we set a maximum number of visible anchor points per-frame to allow efficient batched mapping operations on the GPU.  If all points are matched, it could prevent new ones from being initialized and parts of the scene would lack capacity to adequately represent geometry.  Thus, we perform pruning of matches so that we have a suitable distribution of points before initializing new anchors.  Similar to the GP conditional mean from \cref{eq:gp_cond_mean}, we may also obtain conditional variances of all pixels with respect to the first $j$ pixels via:
\begin{align}
    \boldsymbol{\sigma}_j^2 &= \text{diag}\left[ \Knn \right] - \Knmj \Kmminvj \Kmnj.
\end{align}
We perform active sampling via conditional variance reduction (CVR) \cite{guestrin_near-optimal_2005, moss_inducing_2023}, where at the $(j+1)^\text{th}$ iteration, the pixel in domain $P$ with the largest variance is selected $\pix_{j+1} = \argmax_{\pix \in P} \boldsymbol{\sigma}_j^2 (\pix)$.
We actively sample from the set of potential visible points until a variance threshold is reached, which gives the final set of visible anchor points for the current keyframe.  Compared to \cite{dexheimer_learning_2023}, we found that a minimum distance threshold between pixel locations and from the image border is required to avoid accumulating pixels near the same discontinuity.  This is essential for improving depth estimates and maintaining a good set of candidate points, as otherwise many points near edges may be occluded at the same time.  We show the set of matches between two keyframes in \cref{fig:corr_c}.

\subsection{Anchor Point Initialization}

As the camera explores unobserved parts of the scene, new anchor points must be initialized.  First, we again perform active sampling, but now across all pixels given the current anchor point locations from the previous section.  This ensures new points will be initialized for parts of the scene that are unobserved, lack matches, or require additional geometric capacity, as shown in \cref{fig:corr_d}.

\begin{figure*}[t]
	\centering
	\includegraphics[width=\textwidth]{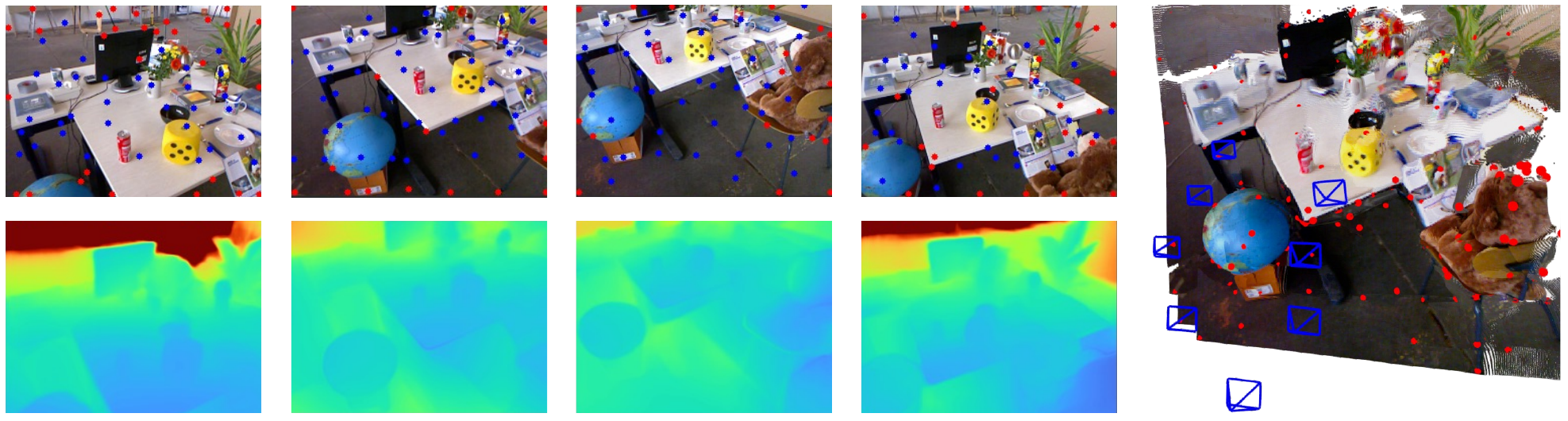}
	\caption{Visualization of sliding-window odometry on TUM \textit{fr2/desk}. Top images show tracked and newly initialized anchor points in blue and red, respectively. Depth maps are below, while the dense point cloud with 248 anchor points is shown on the right.  Note the consistent point tracks on textureless surfaces, such as the table and plate, as well as the long-term tracks on the cube, mug, and book.}
	\label{fig:tum_qual} 
\end{figure*}

Once the pixel locations of new anchor points are known, we initialize the log-depth and backproject into 3D.  Defining the set of known log-depths from the anchor points $\dmone$, the unknown log-depths of newly initialized points $\dmtwo$, and all log-depths as $\dm = (\dmone^T, \dmtwo^T)^T$, we solve another least-squares problem:
\begin{align}
    \label{eq:compress_conditional_depth}
    \min_{\dmtwo}  || \Knm \Kmminv \dm  - \dn ||_{\sigma_d^2} ^2 + ||\dmtwo - s||_{\sigma_s^2},
\end{align}
where the first term promotes fitting reprojected dense depth observations given matched anchor log-depths and the new log-depth initializations, while the second term is a prior that new log-depths are close to the log median depth $s$ of the previous keyframe.  We found that using an isotropic prior with standard deviation $\sigma_d$, rather than $\Kmm$, avoids bias for parts of the image that were not previously observed and would otherwise be underconstrained.  We define $\sigma_d$ as the standard deviation of the residuals from \cref{eq:compress_depth} to better calibrate the model.  Given the new pixel locations, log-depth initializations, and the new keyframe's pose estimate, 3D points are initialized for the backend. A qualitative example of tracked anchor points, depth maps, and reconstruction is shown in \cref{fig:tum_qual}.

\section{Experiments}

We evaluate both the pose and geometry estimation of our method on multiple datasets.  For sparse methods, we compare against feature-based ORB-SLAM3 \cite{campos_orbslam3_2021}, direct DSO \cite{engel_dso_2018}, and learning-based DPVO \cite{teed_dpvo_2023} in trajectory estimation.  We also compare against a representative set of dense methods. TANDEM \cite{koestler_tandem_2022} uses DSO for odometry and fuses learned MVS predictions into a TSDF for tracking.  MonoGS \cite{matsuki_gaussian_2024} is a 3D Gaussian splatting-based monocular SLAM system.  DeepFactors \cite{czarnowski_deepfactors_2020} is a code-based system with photometric, geometric, and keypoint losses to promote consistency across depth maps.  DROID-VO \cite{teed_droid_2021} uses learned correspondence and flow updates for bundle adjustment, such that all depths are reconstructed but lack geometric correlation.  DepthCov \cite{dexheimer_learning_2023} optimizes sparse 2D depths under the depth covariance framework, and in a related manner, COMO-NC (no correspondence) is our full framework but no 3D anchor points are tracked across frames.  We disable loop closure for DeepFactors and DROID-SLAM to isolate the odometry and mapping performance.

\setlength{\tabcolsep}{3pt}
\begin{table}[t]
    \centering
    \scriptsize
    \begin{tabular}{c l|c c c c c c c c|c}
        & & office0 & office1 & office2 & office3 & office4 & room0 & room1 & room2 & mean \\
        \thickhline
        \multirow{2}{*}{S} & ORB-SLAM3 \cite{campos_orbslam3_2021} & 0.4 & 0.3 & 8.5 & 0.4 & 4.5 & 0.3 & 0.3 & 0.4 & 1.9 \\
        & DSO \cite{engel_dso_2018} & 0.3 & 0.5 & 0.3 & 0.2 & 5.3 & 0.2 & 0.8 & 0.2 & \textbf{1.0} \\
        \hline
        \multirow{4}{*}{D} & DeepFactors \cite{czarnowski_deepfactors_2020} & 42.9 & 38.6 & 45.6 & 41.4 & 50.1 & 32.4 & 26.1 & 44.1 & 40.2 \\
        & DROID-VO \cite{teed_droid_2021} & 5.3 & 4.1 & 6.5 & 11.2 & 7.1 & 9.0 & 4.8 & 7.0 & 6.9 \\
        & COMO-NC & 2.9 & 5.0 & 7.4 & 6.1 & 8.1 & 4.4 & 4.5 & 3.7 & 5.3 \\
        & COMO & 2.4 & 2.4 & 4.3 & 3.6 & 5.5 & 2.5 & 2.7 & 3.4 & \textbf{3.4} \\
        \hline
    \end{tabular}
    \caption{Mean ATE (cm) over 5 runs for monocular VO on the Replica dataset. S and D in the first column refer to sparse and dense methods, respectively.}
    \label{tab:replica_ate}
\end{table}

\subsection{Implementation Details}

We use a fixed configuration across all three datasets to demonstrate the robustness of the system.  We use the off-the-shelf depth covariance function \cite{dexheimer_learning_2023} that was trained on the ScanNet training set. Our sliding window odometry uses 9 keyframes and 3 support frames between keyframes for a total of 24.  Each keyframe views a maximum of 64 anchor points and initializes new points if it fails to track all from the previous keyframe.  We use 256x192 images for all operations, and the mapping backend samples pixels with the highest gradient magnitude in 4x4 images patches for photometric error. 

\subsection{Trajectory Evaluation}

\noindent \textbf{Replica} We first test our monocular odometry on the Replica \cite{straub_replica_2019} sequences recorded in \cite{sucar_imap_2021}, which provides a photorealistic environment with ground-truth geometry.  As shown in \cref{tab:replica_ate}, sparse methods without priors perform best, as conditions are favorable for SLAM: few image artifacts, no exposure changes, and sufficient baselines for multi-view geometry.  Among methods with learned priors, COMO achieves the lowest ATE.  The representation has little bias as compared to the code-based DeepFactors, which drifts despite photometric, geometric, and keypoint constraints. Sharing 3D points between frames in COMO shows significant improvement over COMO-NC which disables correspondence. 

\setlength{\tabcolsep}{3pt}
\begin{table}[t]
    \centering
    \scriptsize
    \begin{tabular}{c l|c c c c c c c c c c c|c}
        & & \multicolumn{8}{c}{fr1} & \multicolumn{2}{|c|}{fr2} & \multicolumn{1}{c|}{fr3} & \\
        & & 360 & desk & desk2 & plant & room & rpy & teddy & xyz & \multicolumn{1}{|c}{xyz} & \multicolumn{1}{c|}{desk} & long & mean \\
        \thickhline
        \multirow{3}{*}{S} & ORB-SLAM3 \cite{campos_orbslam3_2021} & X & 2.0 & X & 11.8 & X & 5.6 & X & 1.0 & 0.5 & 1.3 & 1.7 & X \\
        & DSO \cite{engel_dso_2018} & X & 27.2 & 66.0 & 6.0 & 58.6 & X & X & 3.8 & 0.3 & 2.2 & 9.9 & X \\
        & DPVO \cite{teed_dpvo_2023} & 13.1 & 9.4 & 6.5 & 3.0 & 39.8 & 3.5 & 6.2 & 1.3 & 0.5 & 3.5 & 5.5 & \textbf{8.4} \\
        \hline
        \multirow{7}{*}{D} & TANDEM \cite{koestler_tandem_2022} & X & 4.3 & 33.7 & X & X & 4.9 & 43.1 & 2.4 & 0.3 & 2.0 & 8.3 & X \\
        & MonoGS \cite{matsuki_gaussian_2024} & 14.2 & 6.3 & 74.0 & 9.3 & 64.9 & 3.4 & 35.6 & 1.6 & 4.5 & 133.1 & 3.3	& 31.8 \\
        & DeepFactors \cite{czarnowski_deepfactors_2020} & 17.9 & 15.9 & 20.2 & 31.9 & 38.3 & 3.8 & 56.0 & 5.9 & 8.4 & 26.3 & 49.0 & 24.9 \\
        & DepthCov \cite{dexheimer_learning_2023} & 12.8 & 5.6 & 4.8 & 26.1 & 25.7 & 5.2 & 47.5 & 5.6 & 1.2 & 15.9 & 68.8 & 19.9 \\
        & DROID-VO \cite{teed_droid_2021} & 15.7 & 5.2 & 11.1 & 6.0 & 33.4 & 3.2 & 19.1 & 5.6 & 10.7 & 7.9 & 7.3 & 11.4 \\
        & COMO-NC & 16.1 & 4.2 & 10.9 & 19.3 & 28.6 & 5.2 & 68.7 & 4.1 & 0.7 & 8.8 & 46.8 & 19.4 \\
        & COMO & 12.9 & 4.9 & 9.5 & 13.8 & 27.0 & 4.8 & 24.5 & 4.0 & 0.7 & 6.3 & 10.5 & \textbf{10.8} \\
        \hline
    \end{tabular}
    \caption{Mean ATE (cm) over 5 runs for monocular VO on the TUM dataset.  S and D in the first column refer to sparse and dense methods, respectively.}
    \label{tab:tum_ate}
\end{table}

\noindent \textbf{TUM} The TUM RGBD dataset \cite{sturm_benchmark_2012} is a challenging dataset for monocular VO due to significant motion blur, exposure changes, rolling shutter artifacts, and heavy rotations.  Results comparing our method against state-of-the-art VO methods on 11 sequences is shown in \cref{tab:tum_ate}.  For sparse methods, ORB-SLAM3 and DSO both lose tracking on multiple sequences, while DPVO, which uses learned context for tracking 64 sparse patches and performing bundle adjustment, performs best among all methods.  In this work, we are focused on also reconstructing dense geometry, but it would be interesting to combine the learned features of DPVO with our 3D representation.   For dense methods, TANDEM fails due to relying on DSO for odometry, which can still fail despite using a TSDF with MVS fusion for tracking.  This demonstrates the need for joint optimization of poses and dense image information.  While MonoGS can produce accurate trajectories for some sequences, the lack of priors reduces its robustness, and volumetric rendering is slower than all depth-based systems.  Despite promoting consistency between neighboring depth maps, DeepFactors has relatively large error.  Both DepthCov and COMO-NC are similar in terms of optimizing sparse depths in each frame, while leveraging a true 3D representation in COMO reduces error by almost 50\%.  COMO outperforms DROID-VO, which uses learned features for matching and reconstructs all pixels from downsized images via sparse bundle adjustment.  As the assumptions of brightness constancy are violated in many real-world datasets, learned features could further the accuracy of COMO, but it is interesting that our compact representation and simple photometric error can produce the lowest ATE of dense methods.

\begin{figure}[t]
  \begin{minipage}[b]{.7\linewidth}
    \centering
    \includegraphics[width=\linewidth]{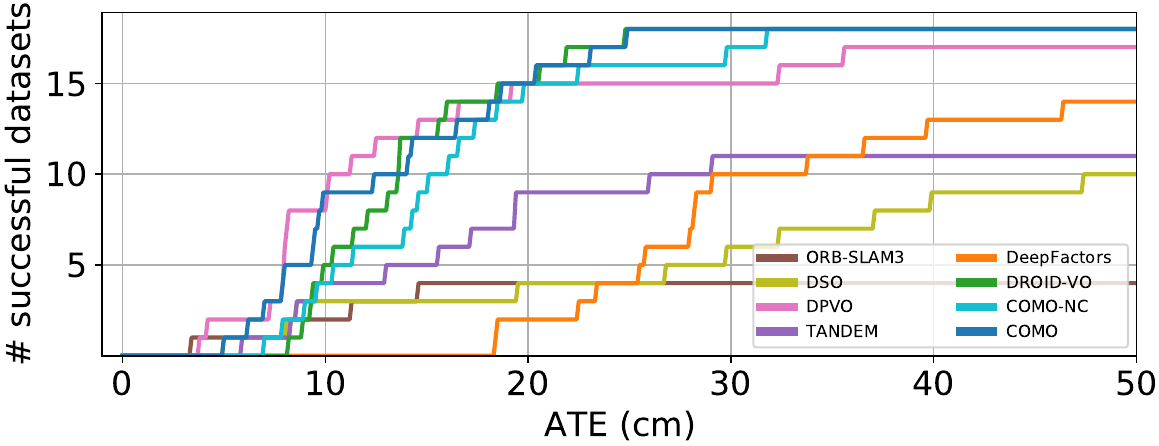}
    \captionsetup{justification=centering}
    \captionof{figure}{Number of successful trajectories below ATE \\ threshold for selected ScanNet test sequences.}
    \label{fig:scannet_traj}
  \end{minipage}\hfill
  \begin{minipage}[b]{.3\linewidth}
    \centering
    \scriptsize
    \begin{tabular}{l|c c}
      Method & ATE & AUC  \\
      \hline
      ORB-SLAM3 & 79.9 & 1.63 \\
      DSO & 54.7 & 2.46 \\
      DPVO & 15.1 & 6.34 \\
      TANDEM & 41.7 & 3.79 \\
      DeepFactors & 36.1 & 2.97 \\
      DROID-VO & 13.9 & 6.51 \\
      COMO-NC & 15.8 & 6.16 \\
      COMO & \textbf{13.0} & \textbf{6.67} \\
    \end{tabular}
    \vspace{18pt}
    \captionsetup{justification=centering}
    \captionof{table}{ScanNet Mean ATE (cm) and AUC.}
    \label{tab:scannet_traj}
  \end{minipage}
\end{figure}

\noindent \textbf{ScanNet Test} Lastly, we evaluate the trajectory error on 18 diverse medium-length scenes from the ScanNet \cite{dai_scannet_2017} test set.  Sequences include homes, offices, schools, businesses, and outdoors.  Methods that fail are assigned a maximum 100 cm error for averages.  Sequences have challenging rotational motion, high image noise, and specular surfaces.  We summarize the results in \cref{fig:scannet_traj}, which shows the number of sequences below varying ATE thresholds, while \cref{tab:scannet_traj} contains summary statistics.  COMO has the lowest mean ATE and area-under-the-curve (AUC).  While DPVO has a higher number of trajectories under low ATE thresholds, it also shows high variance results on some sequences.  Despite using photometric error that is often violated due to image noise and specular surfaces, our representation with 3D consistency proves valuable in challenging real-world data.  An example reconstruction is shown in \cref{fig:title_fig_a}.

\setlength{\tabcolsep}{2pt}

\subsection{Geometry Evaluation}

Since we are interested in both consistent pose estimation and dense geometry, we evaluate dense depth predictions on Replica and ScanNet.  To evaluate accuracy, we perform the same global scale alignment that is used to produce ATE for monocular methods and transform estimated poses and depth maps to be aligned to the metric ground-truth.  Beyond global accuracy, we also measure consistency, which checks whether neighboring frames viewing the same 3D points have consistent depth estimates.  Note that consistency does not care about the correctness of depth.  To check valid pixels, we backproject ground-truth depth maps and project these points into neighboring frames, and check if the depths agree within 1cm.  Then, for all valid 3D correspondences, we calculate metrics between pairs of estimated depth images.  If $N$ is the number of keyframes for that sequence, then $N$ depth images are used for accuracy metrics, while $2(N-1)$ depth images are used for consistency since pairs require both directions.  

\setlength{\tabcolsep}{2pt}
\begin{figure}[t]
  \begin{minipage}[b]{.28\linewidth}
    \centering
    \tiny
    \begin{tabular}{l|c c}
      Method & Replica & ScanNet  \\
      \hline
      TANDEM & N/A & 0.325 \\
      DeepFactors & 0.263 & 0.232 \\
      DROID-VO & 1.129 & 1.389 \\
      COMO-NC & 0.069 & 0.155 \\
      COMO & \textbf{0.046} & \textbf{0.128} \\
    \end{tabular}
    \vspace{20pt}
    \captionsetup{justification=centering}
    \captionof{table}{Depth absolute relative error.}
    \label{tab:depth_eval}
  \end{minipage}
  \hfill
  \begin{minipage}[b]{.71\linewidth}
    \centering
    \includegraphics[width=0.49\linewidth]{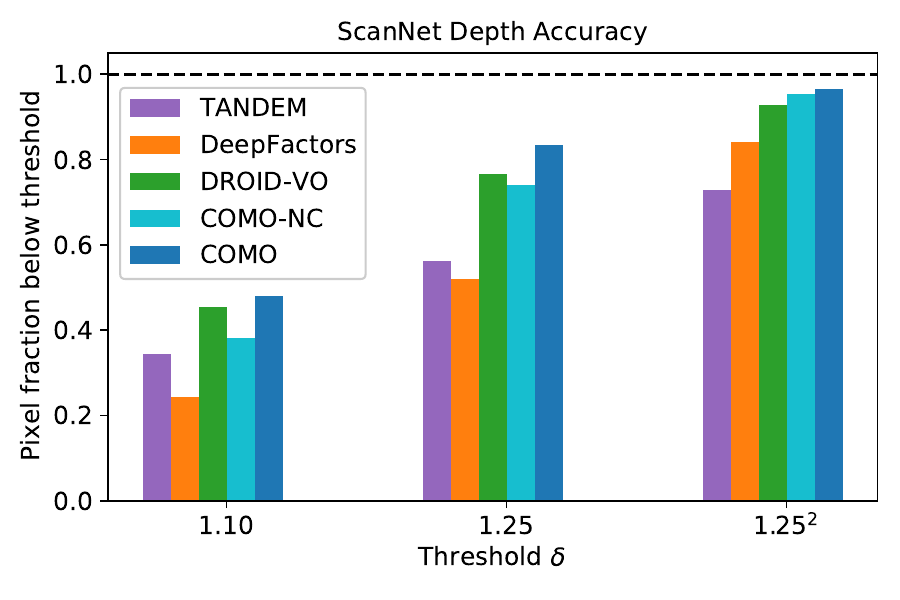}
    \includegraphics[width=0.49\linewidth]{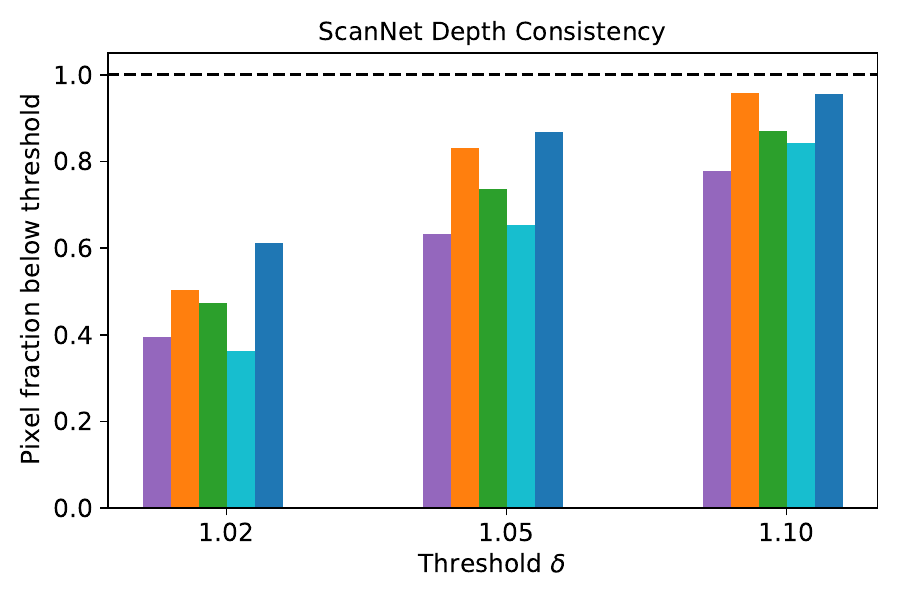}
    \captionsetup{justification=centering}
    \captionof{figure}{ScanNet depth accuracy and consistency metrics for fraction of pixels falling below threshold $\delta$.}
    \label{fig:scannet_depth}
  \end{minipage}
\end{figure}

For estimated depth $\hat{D}$ and ground-truth depth $D$ we show the mean absolute relative error $(1/N) \sum_i |\hat{D}_i-D_i|/D_i$ across all sequences for Replica and ScanNet in \cref{tab:depth_eval}.  We do not evaluate TANDEM on Replica since it is used in the training data.  COMO outperforms other methods due to its ability to jointly optimize dense, correlated geometry along with poses.  Despite accurate pose estimation, DROID-VO has outliers with significant depth errors.  To further evaluate the distribution of depth accuracy and remove the effect of large outliers, in \cref{fig:scannet_depth}, we use the $\delta$ threshold commonly used in depth estimation, which measures what fraction of pixels satisfy $\max{\left(\hat{D}/D, D/\hat{D}\right)} < \delta$.  Interestingly, accurate depth maps do not necessarily imply consistency.  For example, DROID-VO has the second best accuracy behind COMO, while DeepFactors has the second most consistent geometry.  COMO achieves both the best accuracy and consistency without explicit geometric and keypoint constraints.

\begin{figure*}[t]
	\centering
    \begin{subfigure}{0.48\textwidth}
	   \includegraphics[width=\linewidth]{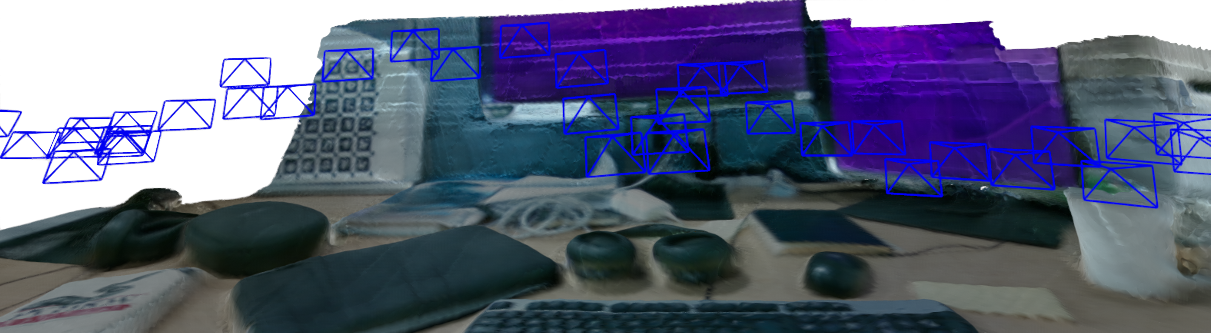}
    \end{subfigure}
    \begin{subfigure}{0.48\textwidth}
	   \includegraphics[width=\linewidth]{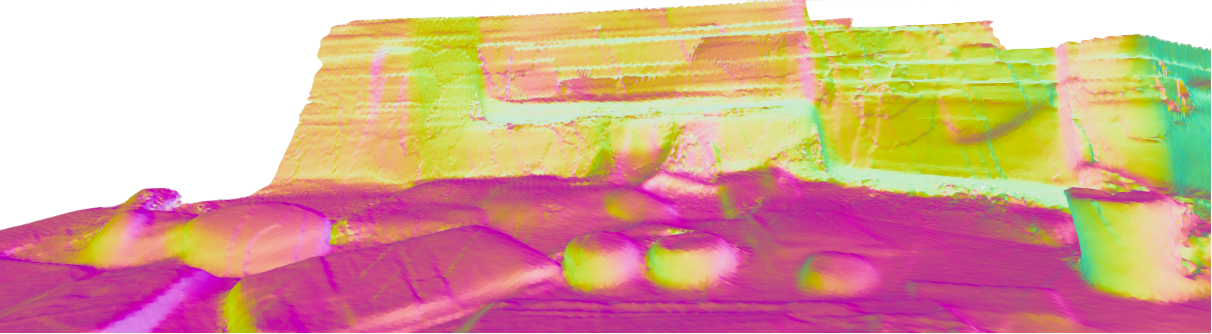}
    \end{subfigure}
    \caption{Mesh and surface normals from TSDF fusion of a live desk sequence.}
    \label{fig:desk_tsdf} 
\end{figure*}

\subsection{Real-time System and Runtime Details} 

When evaluating our system, we use a sequential, single-threaded mode that runs at around 14 FPS with an RTX 3080.  For live demos and reconstructions, we also develop a multiprocessing version that runs at 35-50 FPS as shown in the FPS counter in the top right of the supplementary video, with tracking flexible to be on either CPU or GPU.  We show examples of reconstructions while operating a monocular camera in real-time. \cref{fig:title_fig_b} displays a large-scale raw point cloud of an office, while \cref{fig:desk_tsdf} displays consistent TSDF fusion of a desk sequence. 

Average times for tracking a frame, adding a keyframe (KF), and updating the map are 20.1, 50.4, and 39.8 ms.  Keyframing is the bottleneck during fast motion, but 35-50 FPS in multiprocessing mode for normal motion is standard as shown in the video.  We measure the wall time of several methods in \cref{tab:runtime}.  COMO achieves real-time processing with its efficient representation, but further optimization is still possible since almost all code is implemented in PyTorch.

\setlength{\tabcolsep}{3pt}
\begin{table}[t]
    \centering
    \scriptsize
    \begin{tabular}{c|c c c c c}
          Sequence time & DeepFactors \cite{czarnowski_deepfactors_2020} & MonoGS \cite{matsuki_gaussian_2024} & DROID-VO \cite{teed_droid_2021} & COMO-SP & COMO-MP  \\
          \hline
          % 1m27s & 6m52s & 20m32s & 1m45s & 3m04s & 1m35s
          87 & 412 & 1232 & 105 & 184 & 95
    \end{tabular}
    \caption{Runtime in seconds of different methods on TUM \textit{fr3/long} using an RTX 3080.  COMO SP and MP denote single and multiprocessing, respectively.}
    \label{tab:runtime}
\end{table}

\section{Conclusion}

In this work, we present a compact representation of 3D anchor points and frame-based depth covariance functions to achieve real-time estimation of consistent poses and geometry.  Future work includes training the depth covariance function on more diverse data and replacing photometric constraints with the  more robust learned correspondence from DROID-VO.  Lastly, we believe the anchor point representation could be used in a full map-centric SLAM system with relocalization, which would allow continually improving geometry upon revisiting.  With the advance of learned priors, exploiting geometric correlation is an interesting direction to achieve efficient, robust, and consistent algorithms.  

\section*{Acknowledgements}
This research is funded by EPSRC Prosperity Partnerships (EP/S036636/1) and Dyson Techonology Ltd.  We would like to thank Riku Murai, Hidenobu Matsuki, Gwangbin Bae, and members of the Dyson Robotics Lab for insightful discussions and feedback.

% ---- Bibliography ----
%
% BibTeX users should specify bibliography style 'splncs04'.
% References will then be sorted and formatted in the correct style.
%
\bibliographystyle{splncs04}
\bibliography{bibmap}

% WARNING: Comment this to remove supplementary material at end
\clearpage
\setcounter{page}{1}

\setcounter{section}{0}
\renewcommand{\thesection}{\Alph{section}}

\section{Analytical Jacobians}

One of the main benefits of formulating compact-to-dense geometry in the manner presented is the capability to efficiently calculate analytical Jacobians, which is key to the second-order optimization.  Here, we demonstrate the chain rule required, with each Jacobian being straightforward nonlinear functions that are common in SLAM, such as camera projection and point transformation, linear matrix multiplication from the GP, and element-wise operations, such as exponentiation and logarithm for converting between log-depth and depth.

\subsection{Compact-to-Dense Geometry}

Note that the GP conditional in Equation \ref{eq:gp_cond_mean} means that each dense point $n$ for a given image is dependent on all $m$ anchor points visible to that image.  We show the chain rule for a single dense geometry point $n$ with respect to one visible anchor point $m$ viewed in the keyframe
\begin{align}
   \jac{\Pw^n}{\Pw^m} = \jac{\Pw^n}{\Pck^n} \jac{\Pck^n}{z^n} \jac{z^n}{d^n} \jac{d^n}{\dm} \jac{\dm}{z^m} \jac{z^m}{\Pck^m} \jac{\Pck^m}{\Pw^m},
\end{align}
but note that this can be calculated efficiently by broadcasting and using the full $\Knm \Kmminv$ matrix.  Similarly, we can calculate the Jacobian of a dense geometry point with respect to the keyframe pose it is generated by.  However, note that there are two dependencies on the keyframe pose $\Twck$: one for transforming an anchor point into the camera frame, and the other for transforming a dense geometry point in the camera into the world frame so that other poses may use it in the photometric error calculation.  Therefore, denoting the minimal 6 DoF parameterization of $\mathfrak{se}(3)$ with $\Twc$, and with an abuse of notation of a pose transforming a point as $\Pw = \Twc \Pc$, the Jacobian becomes:
\begin{align}
  \jac{\Pw^n}{\Twck} &= \jac{\left(\Twck \Pck^n \right)}{\Twck} + \jac{\left(\Twck \Pck^n \right)}{\Pck^n} \jac{\Pck^n}{\Twck} \\
  &= \jac{\left(\Twck \Pck^n \right)}{\Twck} \nonumber \\ 
  &+ \jac{\left(\Twck \Pck^n \right)}{\Pck^n} \jac{\Pck^n}{z^n} \jac{z^n}{d^n} \jac{d^n}{\dm} \jac{\dm}{z^m} \jac{z^m}{\Pck^m} \jac{\Pck^m}{\Twck}.
\end{align}

One important design decision was whether to parameterize anchor points as 3D points in the world frame, or depths hosted in a reference camera frame.  While the latter was used in DSO \cite{engel_dso_2018} to reduce the dimensionality of the unknown geometry variables, it would complicate our case.  Using depth hosted in a camera frame includes both the target pose and the reference pose itself in the cost function, and Hessian blocks must now include the reference pose.  In our case, since dense depth maps involve many anchor points, all reference poses associated with corresponding anchor points for a given keyframe would have dependencies.  This would significantly complicate Jacobian calculations, since potentially all poses would be involved in each error computation, rather than just reference keyframe poses and the targets for photometric error.  By keeping the anchor points as world points and projecting, we avoid any host frame dependencies and maintain more efficient and less complex Jacobians.  Furthermore, we also show in Section \ref{subsec:hessian_trick} how to exploit our formulation to reduce the dimensionality of dense Hessian block accumulation.

\subsection{Photometric Residuals}

Image gradients are calculated via finite differences with a 3x3 Scharr filter and bilinearly interpolated along with the grayscale image values.

The Jacobians of the target image value with respect to an anchor point, the reference keyframe pose, and the target image pose are:
\begin{align}
  \jac{\left(I_t\left( \pix^n_t \right) \right)}{\Pw^m} &= \jac{I_t}{\pix^n_t} \jac{ \pix^n_t}{\Pct^n} \jac{\Pct^n} {\Pw^n} \jac{\Pw^n}{\Pw^m}, \\
  \jac{\left(I_t\left( \pix^n_t \right) \right)}{\Twck} &= \jac{I_t}{\pix^n_t} \jac{ \pix^n_t}{\Pct^n} \jac{\Pct^n} {\Pw^n} \jac{\Pw^n}{\Twck}, \\
  \jac{\left(I_t\left( \pix^n_t \right) \right)}{\Twct} &= \jac{I_t}{\pix^n_t} \jac{ \pix^n_t}{\Pct^n} \jac{\Pct^n} {\Twct}.
\end{align}

\subsection{Additional Constraints}

For the two depth priors, we require the Jacobian of the log-depth of each anchor point projection with respect to the anchor point and keyframe pose, which is
\begin{align}
    \jac{d^m}{\Pw^m} &= \jac{d^m}{z^m} \jac{z^m}{\Pck^m} \jac{\Pck^m}{\Pw^m}, \\
    \jac{d^m}{\Twck} &= \jac{d^m}{z^m} \jac{z^m}{\Pck^m} \jac{\Pck^m}{\Twck}.
\end{align}
The pixel prior requires the Jacobian of the pixel projection of the anchor points with repsect to the anchor point and keyframe pose
\begin{align}
    \jac{\pix^m}{\Pw^m} &= \jac{\pix^m}{\Pck^m} \jac{\Pck^m}{\Pw^m}, \\
    \jac{\pix^m}{\Twck} &= \jac{\pix^m}{\Pck^m} \jac{\Pck^m}{\Twck},
\end{align}
which requires the camera projection Jacobian.

\subsection{Geometry Hessian Block Trick}
\label{subsec:hessian_trick}

One of the most expensive steps in the mapping backend is accumulating the Hessian geometry blocks for the photometric error.  For example, with the 64 points per-frame, this is requires a sum of 192x192 matrices over the number of pixels involved in the error.  One interesting component of the geometry Jacobian is the the last term, which has a very simple form:
\begin{align}
    \jac{\Pck^m}{\Pw^m} = \Rwck^T.
\end{align}
Note that for all anchor points in a given keyframe, this Jacobian is identical, as it is just the rotation from world to keyframe coordinates.  Since this is the outer term when accumulating the Hessian, we may factor it out of the sum over all pixels.  Furthermore, the Jacobian for the depth term is also very simple, since it just indexes the z-component of the anchor point in the camera frame:
\begin{align}
    \jac{z^m}{\Pck^m} = [0, 0, 1].
\end{align}
We can also factor this out of the sum, which most importantly, reduces the dimensionality of the Hessian sum to be 64x64 instead of the original 192x192, which greatly improves efficiency by decreasing the memory needed as part of the reduction by 9x.  Furthermore, this reduces pose-geometry blocks from 8x192 to 8x64, which is a 3x reduction. Note that we include each frame's affine brightness parameters in the pose block so it is of dimension 8 instead of 6.

\section{ScanNet Sequences}

 We list the ScanNet test sequences used in Table \ref{tab:scannet_seq}.  We selected these sequences due to the diversity of environments and the fact that they are similar in length.  Some sequences are longer or only rotational motion, which causes all methods to have much larger ATE, at which point it is difficult to tell whether global trajectory measures are useful for evaluating odometry.

\begin{table}[ht]
    \centering
    \begin{tabular}{c|l}
        Sequence & Environment \\ \hline
        709 & Kitchen \\ 
        710 & Home office \\ 
        715 & Reception \\ 
        716 & Laundry \\ 
        719 & Dorm \\ 
        722 & Dorm \\ 
        733 & Living room \\ 
        741 & Bedroom \\ 
        760 & Office \\ 
        780 & Laundry \\ 
        787 & Storage \\ 
        788 & Gym \\
        790 & Copy room \\ 
        792 & Stairs \\
        794 & Outside tables \\ 
        800 & Bookstore \\
        803 & Supply room \\ 
        804 & Copy room \\ 
    \end{tabular}
    \caption{ScanNet test sequences used for evaluation.}
    \label{tab:scannet_seq}
\end{table}

\section{Baseline Details}

To set up DROID-VO from the DROID-SLAM code, we disable the post-process at the end of the sequence that performs global bundle adjustment.  Since we are focused on real-time pose and geometry estimation, this avoids the offline bundle adjustment it performs at the end of a sequence. More specifically, \textit{terminate} function call that first performs two sets of bundle adjustment on all keyframes, and then also the \textit{PoseTrajectoryFiller} that performs motion-only bundle adjustment on not only keyframes, but all frames.  We use the public code for evaluating on TUM, as we found that removing the code that uses every other frame to cause significant errors on \textit{fr1/desk2}, which skews the overall error.  Since DROID-SLAM does not have public code for Replica and ScanNet, we tried different configurations.  For Replica, we found that a resolution of 512x320 to perform better than 320x240, and is closer to the original aspect ratio of the images.  For ScanNet, we found that the images are similar in resolution to TUM, so we use 320x240.  For ScanNet, we must apply a crop of 10 pixels on all sides to handle undistorted regions with invalid values that remain visible.

For DeepFactors, we found that global loop closure could cause large errors, so we disabled it.  As mentioned in the paper, we use all three recommended factors: photometric, reprojection, and geometric.  For Replica, which has a different aspect ratio than TUM and ScanNet, we attempted both cropping and resizing the full image, with the latter giving better overall accuracy.  Since DSO assumes undistorted images, we preprocessed TUM and ScanNet by first undistorting and then cropping any invalid pixels on the edges.  For our method with and without correspondence, we use a 256x192 resolution for all datasets as this resolution is the same as used by the depth covariance UNet.

We compare all methods with only keyframes in the trajectory error, so we extract the up-to date keyframe estimates for evaluation.  For comparing dense depth maps on ScanNet, we performed the same cropping with 10 pixel borders and resizing to a 4:3 ratio for all methods.  This ensures that all predicted depth maps could be registered to the ground-truth depth maps.

\section{Depth Map Comparison}

Since DROID-VO largely treats per-pixel depths independently, this can skew commonly used depth error metrics such as RMSE.  For this reason, we omitted these results from the main paper, but include them here.  We show accuracy and consistency metrics for Replica in Table \ref{tab:depth_replica_accuracy} and Table \ref{tab:depth_replica_consistency}, respectively.  For ScanNet, we show the results in Tables \ref{tab:depth_scannet_accuracy} and \ref{tab:depth_scannet_consistency}.  Bold indicates the best for a given metric, while underline indicates second-best.  Note that for the error metrics (RMSE, MAE, and AbsRel) lower is better, while higher fractions for $\delta$ is better.  COMO achieves the most accurate depths while also being first or second across all consistency metrics.

\setlength{\tabcolsep}{2pt}
\begin{table*}[ht]
    \centering
    \tiny
    \begin{tabular}{l|c c c c c c c c c}
        & RMSE ($\downarrow$) & MAE ($\downarrow$) & AbsRel ($\downarrow$) & $\delta=1.02$ & $\delta=1.05$ & $\delta=1.10$ & $\delta=1.25$ & $\delta=1.25^2$ & $\delta=1.25^3$ \\
        \thickhline
        DeepFactors \cite{czarnowski_deepfactors_2020} & 0.707 & 0.567 & 0.263 & 0.052 & 0.126 & 0.245 & 0.531 & 0.794 & 0.871 \\
        DROID-VO \cite{teed_droid_2021} & 305.131 & 2.534 & 1.129 
        & \underline{0.368} & \underline{0.706} & \underline{0.887} & \underline{0.964} & 0.985 & 0.993 \\
        COMO-NC & \underline{0.256} & \underline{0.160} & \underline{0.069} & 0.228 & 0.513 & 0.773 & 0.963 & \underline{0.992} & \textbf{0.998}  \\
        COMO & \textbf{0.191} & \textbf{0.111} & \textbf{0.046} & \textbf{0.371} & \textbf{0.716} & \textbf{0.895} & \textbf{0.976} & 	\textbf{0.994} & \textbf{0.998}
    \end{tabular}
    \caption{Depth accuracy evaluation on Replica.}
    \label{tab:depth_replica_accuracy}
\end{table*}

\setlength{\tabcolsep}{2pt}
\begin{table*}[ht]
    \centering
    \tiny
    \begin{tabular}{l|c c c c c c c c c}
        & RMSE ($\downarrow$) & MAE ($\downarrow$) & AbsRel ($\downarrow$) & $\delta=1.02$ & $\delta=1.05$ & $\delta=1.10$ & $\delta=1.25$ & $\delta=1.25^2$ & $\delta=1.25^3$ \\
        \thickhline
        DeepFactors \cite{czarnowski_deepfactors_2020} & \textbf{0.097} & \textbf{0.057} & \underline{0.029} & 0.500 &	0.845 &	\textbf{0.968} &	\textbf{0.997} &	\textbf{0.999} &	\textbf{0.999} \\
        DROID-VO \cite{teed_droid_2021} & 258.511 &	1.689 &	0.248 &	\textbf{0.729} &	\textbf{0.901} &	0.953 &	0.976 &	0.982 &	0.985 \\
        COMO-NC & 0.181 & 0.098 &	0.040 &	0.455 &	0.760 &	0.912 &	0.984 &	0.997 &	\textbf{0.999}  \\
        COMO & \underline{0.119} &	\underline{0.060} &	\textbf{0.025} &	\underline{0.630} &	\underline{0.879} &	\underline{0.962} &	\underline{0.993} &	\underline{0.998} &	\textbf{0.999}
    \end{tabular}
    \caption{Depth consistency evaluation on Replica.}
    \label{tab:depth_replica_consistency}
\end{table*}

\setlength{\tabcolsep}{2pt}
\begin{table*}[ht]
    \centering
    \tiny
    \begin{tabular}{l|c c c c c c c c c}
        & RMSE ($\downarrow$) & MAE ($\downarrow$) & AbsRel ($\downarrow$) & $\delta=1.02$ & $\delta=1.05$ & $\delta=1.10$ & $\delta=1.25$ & $\delta=1.25^2$ & $\delta=1.25^3$ \\
        \thickhline
        TANDEM \cite{koestler_tandem_2022} & 0.923 & 0.655	& 0.325	& 0.084	& 0.193	& 0.343	& 0.562	& 0.729	& 0.793 \\
        DeepFactors \cite{czarnowski_deepfactors_2020} & 0.684	& 0.499	& 0.232	& 0.053	& 0.128	& 0.243	& 0.519	& 0.842	& 0.959 \\
        DROID-VO \cite{teed_droid_2021} & 481.428 & 3.138 & 1.389 & \underline{0.114} & \underline{0.264} & \underline{0.454} & \underline{0.766} & 0.928 & 0.964 \\
        COMO-NC & \underline{0.464} & \underline{0.332} & \underline{0.155} & 0.086 & 0.210 & 0.381 & 0.741 & \underline{0.955} & \textbf{0.990}  \\
        COMO & \textbf{0.416} & \textbf{0.278} & \textbf{0.128} & \textbf{0.123} & \textbf{0.281} & \textbf{0.480} & \textbf{0.835} & \textbf{0.965} & \underline{0.987}
    \end{tabular}
    \caption{Depth accuracy evaluation on ScanNet.}
    \label{tab:depth_scannet_accuracy}
\end{table*}

\setlength{\tabcolsep}{2pt}
\begin{table*}[ht]
    \centering
    \tiny
    \begin{tabular}{l|c c c c c c c c c}
        & RMSE ($\downarrow$) & MAE ($\downarrow$) & AbsRel ($\downarrow$)  & $\delta=1.02$ & $\delta=1.05$ & $\delta=1.10$ & $\delta=1.25$ & $\delta=1.25^2$ & $\delta=1.25^3$ \\
        \thickhline
        TANDEM \cite{koestler_tandem_2022} & 0.366 & 0.150 & 0.160 & 	0.395 & 0.632 & 0.778 & 0.891 & 0.938 & 0.956 \\
        DeepFactors \cite{czarnowski_deepfactors_2020} & \textbf{0.097} & \textbf{0.051} & \underline{0.031} & \underline{0.503} & \underline{0.831} & \textbf{0.958} & \textbf{0.996} & \textbf{0.998} & \textbf{0.998} \\
        DROID-VO \cite{teed_droid_2021} & 244.693 & 1.140 & 0.258 &	0.474 & 0.736 & 0.869 & 0.947 & 0.969 & 0.975 \\
        COMO-NC & 0.201 & 0.106 & 0.056 & 0.363 & 0.654 & 0.842 & 0.965 & 0.993 & 0.997  \\
        COMO & \underline{0.114} & \textbf{0.051} & \textbf{0.028} & \textbf{0.612} & \textbf{0.867} & \underline{0.955} & \underline{0.991} & \underline{0.997} & \textbf{0.998}
    \end{tabular}
    \caption{Depth consistency evaluation on ScanNet.}
    \label{tab:depth_scannet_consistency}
\end{table*}

\end{document}